\PassOptionsToPackage{unicode}{hyperref}
\PassOptionsToPackage{hyphens}{url}
\PassOptionsToPackage{dvipsnames,svgnames,x11names}{xcolor}
\documentclass[
  11pt,
]{article}
\usepackage{amsmath,amssymb}
\usepackage{iftex}
\ifPDFTeX
  \usepackage[T1]{fontenc}
  \usepackage[utf8]{inputenc}
  \usepackage{textcomp} 
\else 
  \usepackage{unicode-math} 
  \defaultfontfeatures{Scale=MatchLowercase}
  \defaultfontfeatures[\rmfamily]{Ligatures=TeX,Scale=1}
\fi
\usepackage{lmodern}
\ifPDFTeX\else
\fi
\IfFileExists{upquote.sty}{\usepackage{upquote}}{}
\IfFileExists{microtype.sty}{
  \usepackage[]{microtype}
  \UseMicrotypeSet[protrusion]{basicmath} 
}{}
\makeatletter
\@ifundefined{KOMAClassName}{
  \IfFileExists{parskip.sty}{%
    \usepackage{parskip}
  }{
    \setlength{\parindent}{0pt}
    \setlength{\parskip}{6pt plus 2pt minus 1pt}}
}{
  \KOMAoptions{parskip=half}}
\makeatother
\usepackage{xcolor}
\usepackage[margin=1in]{geometry}
\usepackage{longtable,booktabs,array}
\usepackage{calc} 
\usepackage{etoolbox}
\makeatletter
\patchcmd\longtable{\par}{\if@noskipsec\mbox{}\fi\par}{}{}
\makeatother
\IfFileExists{footnotehyper.sty}{\usepackage{footnotehyper}}{\usepackage{footnote}}
\makesavenoteenv{longtable}
\usepackage{graphicx}
\makeatletter
\def\maxwidth{\ifdim\Gin@nat@width>\linewidth\linewidth\else\Gin@nat@width\fi}
\def\maxheight{\ifdim\Gin@nat@height>\textheight\textheight\else\Gin@nat@height\fi}
\makeatother
\setkeys{Gin}{width=\maxwidth,height=\maxheight,keepaspectratio}
\makeatletter
\def\fps@figure{htbp}
\makeatother
\usepackage{graphicx}
\setkeys{Gin}{width=\linewidth,height=0.92\textheight,keepaspectratio}
\graphicspath{{./}}
\usepackage{booktabs,longtable,array,multirow}
\usepackage{etoolbox}
\AtBeginEnvironment{longtable}{\scriptsize\setlength{\tabcolsep}{2pt}%
  \renewcommand{\_}{\textunderscore\allowbreak}}
\usepackage{pdflscape}   
\usepackage[htt]{hyphenat}   
\AtBeginDocument{\renewcommand{\_}{\textunderscore\allowbreak}}
\usepackage{microtype}
\usepackage{xurl}                 
\usepackage{newunicodechar}
\newunicodechar{⁰}{\textsuperscript{0}}\newunicodechar{¹}{\textsuperscript{1}}
\newunicodechar{²}{\textsuperscript{2}}\newunicodechar{³}{\textsuperscript{3}}
\newunicodechar{⁴}{\textsuperscript{4}}\newunicodechar{⁵}{\textsuperscript{5}}
\newunicodechar{⁶}{\textsuperscript{6}}\newunicodechar{⁷}{\textsuperscript{7}}
\newunicodechar{⁸}{\textsuperscript{8}}\newunicodechar{⁹}{\textsuperscript{9}}
\newunicodechar{⁺}{\textsuperscript{+}}\newunicodechar{⁻}{\textsuperscript{-}}
\newunicodechar{⁼}{\textsuperscript{=}}\newunicodechar{ⁿ}{\textsuperscript{n}}
\newunicodechar{₀}{\textsubscript{0}}\newunicodechar{₁}{\textsubscript{1}}
\newunicodechar{₂}{\textsubscript{2}}\newunicodechar{₃}{\textsubscript{3}}
\newunicodechar{₄}{\textsubscript{4}}\newunicodechar{₅}{\textsubscript{5}}
\newunicodechar{₆}{\textsubscript{6}}\newunicodechar{₇}{\textsubscript{7}}
\newunicodechar{₈}{\textsubscript{8}}\newunicodechar{₉}{\textsubscript{9}}
\newunicodechar{₊}{\textsubscript{+}}\newunicodechar{₋}{\textsubscript{-}}
\newunicodechar{·}{\textperiodcentered}
\newunicodechar{≥}{\ensuremath{\geq}}\newunicodechar{≤}{\ensuremath{\leq}}
\newunicodechar{×}{\ensuremath{\times}}\newunicodechar{±}{\ensuremath{\pm}}
\newunicodechar{≈}{\ensuremath{\approx}}\newunicodechar{≠}{\ensuremath{\neq}}
\newunicodechar{∼}{\ensuremath{\sim}}\newunicodechar{−}{\ensuremath{-}}
\newunicodechar{→}{\ensuremath{\rightarrow}}\newunicodechar{←}{\ensuremath{\leftarrow}}
\newunicodechar{✓}{\ensuremath{\checkmark}}\newunicodechar{✗}{\ensuremath{\times}}
\newunicodechar{∆}{\ensuremath{\Delta}}\newunicodechar{Δ}{\ensuremath{\Delta}}
\newunicodechar{α}{\ensuremath{\alpha}}\newunicodechar{β}{\ensuremath{\beta}}
\newunicodechar{γ}{\ensuremath{\gamma}}\newunicodechar{δ}{\ensuremath{\delta}}
\newunicodechar{ε}{\ensuremath{\varepsilon}}\newunicodechar{θ}{\ensuremath{\theta}}
\newunicodechar{λ}{\ensuremath{\lambda}}\newunicodechar{μ}{\ensuremath{\mu}}
\newunicodechar{π}{\ensuremath{\pi}}\newunicodechar{ρ}{\ensuremath{\rho}}
\newunicodechar{σ}{\ensuremath{\sigma}}\newunicodechar{τ}{\ensuremath{\tau}}
\newunicodechar{φ}{\ensuremath{\varphi}}\newunicodechar{χ}{\ensuremath{\chi}}
\newunicodechar{ψ}{\ensuremath{\psi}}\newunicodechar{ω}{\ensuremath{\omega}}
\newunicodechar{Σ}{\ensuremath{\Sigma}}\newunicodechar{Ω}{\ensuremath{\Omega}}
\newunicodechar{ᵢ}{\textsubscript{i}}\newunicodechar{ₐ}{\textsubscript{a}}
\newunicodechar{ᵇ}{\textsuperscript{b}}
\newunicodechar{↔}{\ensuremath{\leftrightarrow}}
\newunicodechar{☆}{\ensuremath{\star}}\newunicodechar{★}{\ensuremath{\bigstar}}
\newunicodechar{⌈}{\ensuremath{\lceil}}\newunicodechar{⌉}{\ensuremath{\rceil}}
\newunicodechar{⌊}{\ensuremath{\lfloor}}\newunicodechar{⌋}{\ensuremath{\rfloor}}
\newunicodechar{′}{\ensuremath{{}^{\prime}}}\newunicodechar{″}{\ensuremath{{}^{\prime\prime}}}
\newunicodechar{𝒩}{\ensuremath{\mathcal{N}}}
\usepackage{titlesec}
\titleformat{\section}{\large\bfseries}{}{0pt}{}
\titleformat{\subsection}{\normalsize\bfseries}{}{0pt}{}
\titleformat{\subsubsection}{\normalsize\itshape\bfseries}{}{0pt}{}
\titlespacing*{\section}{0pt}{1.3em}{0.45em}
\titlespacing*{\subsection}{0pt}{1.0em}{0.35em}
\titlespacing*{\subsubsection}{0pt}{0.8em}{0.3em}
\newcommand{\TitleBlock}[5]{%
  \begin{center}
  {\LARGE\bfseries #1\par}
  \ifx\relax#2\relax\else\vspace{0.45em}{\large\itshape #2\par}\fi
  \vspace{0.9em}{\normalsize #3\par}
  \vspace{0.5em}{\footnotesize #4\par}
  \ifx\relax#5\relax\else\vspace{0.4em}{\footnotesize #5\par}\fi
  \end{center}\vspace{0.6em}\hrule\vspace{1.4em}}
\newunicodechar{ }{\,}
\newunicodechar{ }{\,}
\newunicodechar{ }{\,}
\newunicodechar{ }{\,}
\newunicodechar{ }{~}
\newunicodechar{ }{\,}
\ifLuaTeX
  \usepackage{selnolig}  
\fi
\IfFileExists{bookmark.sty}{\usepackage{bookmark}}{\usepackage{hyperref}}
\IfFileExists{xurl.sty}{\usepackage{xurl}}{} 
\hypersetup{
  colorlinks=true,
  linkcolor={Maroon},
  filecolor={Maroon},
  citecolor={Blue},
  urlcolor={Blue},
  pdfcreator={LaTeX via pandoc}}

\author{}
\date{}

\begin{document}

\textbf{Perturbation-based Regional Interpretability through Subtraction
Mapping (PRISM): naming-error dissociations in language models and
post-stroke aphasia}

Xiang Guan¹†, Roger D. Newman-Norlund²,³†*, Yong Yang¹, Saeed Ahmadi²,
Regan Willis¹, Nadra Salman⁴, Kalil Warren⁴, Srihari Nelakuditi¹, Chris
Rorden⁵, Leonardo Bonilha⁶, Julius Fridriksson²,³

¹ Department of Computer Science and Engineering, University of South
Carolina ² Department of Communication Sciences and Disorders,
University of South Carolina ³ ALLT.AI, LLC, Columbia, SC ⁴ Linguistics
Program, University of South Carolina ⁵ Department of Psychology,
University of South Carolina ⁶ Department of Neurology, USC School of
Medicine

† Xiang Guan and Roger D. Newman-Norlund contributed equally to this
work and are joint first authors.

*Corresponding author: Roger D. Newman-Norlund (rnorlund@mailbox.sc.edu)

\textbf{Abstract}

Mechanistic interpretability of large language models lacks spatially
resolved, falsifiable tools for testing whether internal components are
specialized for distinct cognitive operations. We adapt subtraction
analysis, the standard framework of human neuroimaging, from biological
brains to perturbed transformers, and apply the same logic to both
substrates in parallel. Building on the Brain-LLM Unified Model (BLUM),
which showed that layer-perturbed LLaVA-1.6-Vicuna-13B error profiles
match the lesion patterns of aphasic patients, we develop PRISM
(Perturbation-based Regional Interpretability through Subtraction
Mapping). PRISM maps the seven clinical Philadelphia Naming Test
categories, subtracts error classes pairwise, and treats each
perturbation seed as a subject in a group analysis with threshold-free
cluster enhancement along the layer axis. We run a structurally matched
analysis on 213 chronic post-stroke aphasia patients using
correlation-difference lesion-symptom mapping, and replicate both sides
on held-out splits. The designs match in subject dimension (seeds,
patients), spatial dimension (layers, atlas-parcellated cortex) and
thresholding, but the contrast operator differs: a within-subject
error-proportion difference for the LLM, a between-subject correlation
difference for the cortex. Both substrates recover a robust
phonemic-favoring dissociation, a deep layer cluster and a
frontal-perisylvian cortical cluster, both replicating; the
semantic-favoring direction is a consistently signed but non-significant
trend on both. PRISM thus gives a falsifiable, spatially resolved test
of functional-specialization claims in transformer language models. A
confirmatory ROI-level intervention (PRISM Stage 3) licensing the
strongest causal-mechanism claim is left to subsequent work.


\textbf{Introduction}

\textbf{The interpretability gap}

Mechanistic interpretability of large language models (LLMs) is
dominated by methods that operate on internal representations or
weights: probing classifiers (Belinkov 2022; Hewitt and Manning 2019),
causal tracing and weight editing (Meng et al.~2022; Geiger et al.~2021;
Wu et al.~2023), automated circuit discovery (Conmy et al.~2023), and
sparse-autoencoder feature decomposition (Cunningham et al.~2023;
Templeton et al.~2024). Each yields evidence about which components
encode or compute what, but each evaluates its claims against benchmarks
or against internal reconstruction quality, not against an externally
validated, behaviorally defined map of what the model's components are
causally necessary for. The result is a fragmented landscape of
interpretability claims with no shared inferential framework for testing
whether components are functionally specialized for distinct cognitive
operations.

The neighboring field of human cognitive neuroscience has spent decades
developing exactly such a framework. Subtraction analysis, applied to
functional neuroimaging maps (Petersen et al.~1988; Friston et al.~1995)
and to lesion--symptom maps (Bates et al.~2003; Mirman et al.~2015)
under nonparametric multiple-comparison correction (Smith and Nichols
2009; Maris and Oostenveld 2007), has parceled the human brain into
regions whose involvement is selective to one operation versus another.
Replication and targeted intervention close the inferential loop. The
framework is informative when the loci of interest are spatially
organized within a common domain, and the operations being contrasted
are approximately separable. Here, we ask what happens when those tools
are imported into LLM interpretability, and we run them in parallel on
the LLM and on a cohort of chronic-aphasia patients administered the
same behavioral task.

\textbf{Perturbation-induced error maps and PRISM}

A growing body of work has begun to bridge LLM and brain analysis by
treating LLM activations as data structurally analogous to neural
recordings (Schrimpf et al.~2021; Caucheteux and King 2022; Goldstein et
al.~2022, 2025; Antonello et al.~2024; Tuckute et al.~2024; Kumar et
al.~2024; Mischler et al.~2025; Gao et al.~2025). The Brain--LLM Unified
Model (BLUM; Fridriksson et al.~2026) extended this analogy to
behavioral output. By administering the Philadelphia Naming Test (PNT)
and the Western Aphasia Battery--Revised (WAB-R) to a perturbed
13-billion-parameter vision--language transformer (LLaVA-1.6-Vicuna-13B;
Liu et al.~2023; Chiang et al.~2023) and to humans with chronic
post-stroke aphasia (N = 410), an impairment in language processing, and
by projecting the resulting LLM error profiles through human-trained
symptom-to-lesion models, BLUM showed that LLM-derived predicted lesions
corresponded to actual lesions in error-matched humans above chance in
67\% of picture-naming (p \textless{} 10⁻⁵³) and 68\% of
sentence-completion conditions (p \textless{} 10⁻⁶⁸). The structural
implication is direct: systematic perturbation of an LLM produces error
data of the same form that lesion--symptom mapping has analyzed in
humans for over a century: an ordered spatial domain (the transformer's
layers; Tenney et al.~2019; Geva et al.~2021; Rogers et al.~2020),
behavioral consequences classified into a clinically validated taxonomy
(Schwartz et al.~2006; Dell et al.~1997), and replicable layerwise
variation.

Two features distinguish the artificial setting from cortex. First,
residual connections couple every layer's output to the activations of
all preceding layers (Elhage et al.~2021), so a layer-targeted
perturbation may not isolate that layer's computation as cleanly as a
focal lesion isolates a cortical region. Second, every component of the
analysis can be repeated at will, with independent random perturbation
seeds for replication and with the relevant lesion delivered or withheld
at the experimenter's choice. This combination of structural similarity
to lesion--symptom data and the lack of controllability in biological
systems motivates applying the lesion-symptom-analogous analytic
pipeline to LLM perturbation maps. For the sake of brevity, we call the
resulting framework PRISM (Perturbation-based Regional Interpretability
through Subtraction Mapping). The technical lineage imports the
subtraction logic from fMRI activation mapping (Petersen et al.~1988;
Friston et al.~1995), together with cluster-based spatial inference
originally developed for fMRI (Smith and Nichols 2009; Maris and
Oostenveld 2007), into a perturbation design that is structurally
analogous to lesion--symptom mapping (Bates et al.~2003; Mirman et
al.~2015), an analogy licensed by BLUM's demonstration (Fridriksson et
al.~2026) that LLM perturbation profiles map onto real human lesion
patterns. The contrast logic, the cross-replicate inferential machinery,
and the cluster-correction step are inherited from the
activation-mapping tradition; the perturbation-as-lesion framing, the
necessity-of-layer-cluster reading of significant regions, and the
confirmatory-intervention closing step are inherited from the
lesion-symptom tradition. In the present paper, we run subtraction
analyses in parallel on both substrates so that the cross-substrate
correspondence is demonstrated within the study rather than asserted by
analogy.

PRISM is complementary to the dominant tradition in LLM mechanistic
interpretability rather than competing with it. Activation patching and
causal mediation analysis (Vig et al.~2020; Meng et al.~2022; Geiger et
al.~2021; Wu et al.~2023; Heimersheim and Nanda 2024; Zhang and Nanda
2024), circuit-level analysis and discovery (Wang et al.~2023; Conmy et
al.~2023; Merullo et al.~2024), and sparse-autoencoder feature
decomposition (Cunningham et al.~2023; Templeton et al.~2024) operate on
internal representations or weights and yield fine-grained
component-level claims. PRISM, by contrast, is phenotype-first and
intentionally coarse-grained: it begins from a clinically defined
behavioral failure, a ground truth in the form of an aphasic error
category, and asks which layer ranges are causally implicated in
producing it, working at the level of layer clusters rather than
individual heads or features. The two approaches answer complementary
questions: internal-representation methods identify which components
carry or compute which information; PRISM identifies which layer ranges
are causally necessary to prevent which behavioral failures. The coarse
granularity is a feature in this regime, not a concession. It is what
supplies behavioral falsifiability, statistical power, and the
multiple-comparison and confirmatory-intervention machinery that have,
in human cognitive neuroscience, made such inferences trustworthy.

\textbf{The PRISM pipeline}

For each of the six clinical PNT error categories (semantic, phonemic,
mixed, neologism, no response, and unrelated), PRISM constructs a
perturbation-induced layer map across all 40 transformer layers and a
parallel cortical map across the human patient cohort. The framework
then computes pairwise subtractions between error categories, applies
threshold-free cluster enhancement (TFCE) for spatial inference, and
replicates the resulting maps on an independent split (held-out
perturbation seeds for the LLM; held-out patients for the cortex).
Because layer adjacency does not guarantee functional adjacency, we
additionally validate TFCE on the LLM side under permutation of layer
order: true clusters should dissolve when the layer sequence is
randomized. We further test the separability assumption underlying
subtraction logic by repeating the analysis under random permutation of
error-category labels, where chance-level subtraction maps would
indicate that the operations are not separable in this model. Each
component of this pipeline has a long history in human cognitive
neuroscience; what is new is its application to LLM-derived behavioral
error maps under the same machinery used on the human cortex within this
paper.

The question subtraction was designed to answer translates directly to
the perturbed LLM: which layer clusters, when perturbed, preferentially
increase one error type over another? Because the relevant lesion can be
created at will, candidate clusters identified by this exploratory
analysis are confirmed by targeted region-of-interest (ROI)
perturbation: perturbing only the implicated layers, perturbing matched
non-significant control clusters as a null, and verifying that the
resulting error profiles match the maps' predictions selectively for the
implicated clusters. PRISM thus consists of three sequential stages:
exploratory subtraction analysis with TFCE and multi-seed replication;
layer-order- and label-permutation validation of TFCE assumptions; and
confirmatory ROI perturbation against null controls.

In the present study, we apply PRISM to the same vision--language
transformer evaluated in BLUM (LLaVA-1.6-Vicuna-13B; 40 transformer
layers, 13 billion parameters) and to the Philadelphia Naming Test
(Roach et al.~1996), the gold-standard task for measuring anomia, the
hallmark language impairment in aphasia (Goodglass and Kaplan 1983;
Laine and Martin 2023). The PNT was selected for its 175-item depth,
six-category error taxonomy, and deep neurolinguistic and
neuropsychological development, which together provide the behavioral
signal-to-noise required to estimate stable maps for each error category
and to compute pairwise subtractions across categories. In parallel, the
same confirmatory subtraction analyses are run on the BLUM patient
cohort under standard clinical conditions (276 administered the PNT; 213
with complete JHU lesion-load profiles enter the analysis), so that the
LLM and human pipelines are matched in subject dimension, spatial
dimension, and TFCE-style thresholding, even where the contrast operator
differs at the technical level. Restricting to a single task and a
single architecture is a deliberate scope decision; extending to the
WAB-R and to text-only or encoder--decoder architectures is a defined
next step. Subtraction maps and, on the LLM side, ROI confirmation
profiles for each naming error category are the outputs of the
framework.

PRISM, as developed here, supplies a behaviorally grounded test of which
layer ranges are necessary for which error categories. The pipeline is
portable across architectures, tasks, and behavioral taxonomies; the
inferential standards are inherited unchanged from human cognitive
neuroscience; and the confirmatory ROI step closes the
inference-versus-correlation loop in a way that lesion--symptom mapping
in human patients structurally cannot. The framework supports two
ultimate goals that together define the broader research program
(Fridriksson et al.~2026). The first is improving the explainability of
large language models, by providing interpretability evidence of a kind
that internal-representation methods alone cannot deliver, framed in an
external, clinically validated taxonomy of language failure accessible
to clinicians and to the AI community alike. The second is the
construction of patient-specific LLM digital twins of neurological
disorders. By identifying the perturbation conditions whose error
profiles most closely reproduce those of an individual patient, and by
manipulating those perturbations under the inferential standards PRISM
establishes, computational surrogates can be constructed that simulate
patient-specific deficits, model disease trajectories, and predict
treatment responses, with direct application to clinical-trial design
and acceleration in conditions where recruitment and longitudinal
follow-up are rate-limiting.

\textbf{Methods}

\textbf{Behavioral readout: the Philadelphia Naming Test}

We use the Philadelphia Naming Test (PNT; Roach et al.~1996), a 175-item
picture-naming task that elicits single-word responses. Each response is
scored under the standard PNT clinical taxonomy (Schwartz et al.~2006;
Dell et al.~1997) into one Correct category plus six error categories,
for seven categories total (Table 1). The same taxonomy is applied to
LLM responses and to human patient responses, so that the input to
subtraction analysis is comparable across substrates.

On the LLM side, the analysis is restricted to the items the unperturbed
model already names correctly. The full 175-item PNT was administered to
LLaVA-1.6-Vicuna-13B without perturbation, and the 158 items it named
correctly at baseline define the analysis set; all perturbation
conditions are scored on those 158 items. The rationale is that an item
the model fails at baseline cannot be informative about the effect of a
perturbation: its error is a property of the unperturbed model rather
than of the manipulation, so retaining it would add a fixed error floor
that is common to every cell of the perturbation grid and dilute every
contrast equally. Restricting to baseline-correct items makes every
error observed in the analysis attributable to the perturbation.
Per-cell proportions are computed against the number of items actually
scored in that cell, which is 158 except in 18 of 317,346
administrations where a single response could not be scored and the
denominator is 157. The human-side analysis is not restricted in this
way, since the patient cohort has no analogous baseline-correct
condition.

\begin{longtable}[]{@{}
  >{\raggedright\arraybackslash}p{(\columnwidth - 2\tabcolsep) * \real{0.5000}}
  >{\raggedright\arraybackslash}p{(\columnwidth - 2\tabcolsep) * \real{0.5000}}@{}}
\toprule\noalign{}
\begin{minipage}[b]{\linewidth}\raggedright
Category
\end{minipage} & \begin{minipage}[b]{\linewidth}\raggedright
Definition
\end{minipage} \\
\midrule\noalign{}
\endhead
\bottomrule\noalign{}
\endlastfoot
\textbf{Correct} & The target word is produced (e.g., ``cat'' for the
image of a cat). \\
\textbf{Semantic substitution} & A real word that is semantically
related to the target (e.g., ``dog'' for cat). \\
\textbf{Phonemic} & An attempt that shares phonological form with the
target but is either a real word with one or more sound substitutions (a
formal paraphasia, e.g., ``hat'' for cat) or a non-word that closely
resembles the target's sound shape (e.g., ``kag'' for cat). Following
standard practice we merge the formal and nonword sub-categories into a
single Phonemic category. \\
\textbf{Mixed} & A real word related to the target in both meaning and
sound (e.g., ``rat'' for cat). \\
\textbf{Neologism} & A non-word string that does not closely resemble
the target's sound shape (e.g., ``blorf'' for cat). \\
\textbf{Unrelated} & A real word with no clear semantic or phonological
relation to the target (e.g., ``wheel'' for cat). \\
\textbf{NoResponse} & Failure to produce any scorable utterance within
the response window. \\
\end{longtable}

\textbf{Table 1.} The seven response categories of the Philadelphia
Naming Test (PNT; Roach et al.~1996), scored under the standard clinical
taxonomy of Schwartz et al.~(2006) and Dell et al.~(1997). Each row
gives the category name, its definition, and a worked example for the
target image of a cat. The Phonemic category collapses the
formal-paraphasia and nonword sub-categories into a single class
following standard practice. Correct is held out of pairwise contrasts
in PRISM because contrasts against Correct re-express overall error rate
rather than category-selective involvement; the six error categories
below the Correct row are the inputs to pairwise subtraction analysis.

\textbf{Patient lesion data}

The BLUM patient cohort (Fridriksson et al.~2026) was administered the
PNT under standard clinical conditions. All study procedures were
approved by the University of South Carolina Institutional Review Board
(Pro00053559) and all participants provided written informed consent;
the data analyzed here are archival, and no new human data were
collected for the present study. Of the 276 patients administered,
\(N = 213\) had both a valid PNT administration and a complete JHU ROI
lesion-load profile and entered the analyses reported here; the
remaining 63 were excluded prior to any analysis at the lesion-mask
quality-control step. Each patient contributes a vector of PNT
error-category proportions and a 3D binary lesion mask in MNI152 space
at 1 mm isotropic resolution (lesion masks were traced from structural
MRI by trained raters). Per-patient lesion load is summarized at the
level of the Johns Hopkins University (JHU) white-matter and grey-matter
atlas (Mori et al.~2008) by computing, for each atlas region, the
proportion of its voxels that lie within the patient's binary lesion
mask. We retain the 64 left-hemisphere regions surviving a 10\% damage
threshold (i.e., at least some patient has \(\geq 10\%\) of the region
lesioned) with ventricles dropped.

\textbf{LLM perturbation grid}

The artificial-network analogue of a focal cortical lesion is delivered
by directly perturbing the weights of a single transformer layer of
LLaVA-1.6-Vicuna-13B (Liu et al.~2023; Chiang et al.~2023), one of 40
successive processing stages in the model, with multiplicative Gaussian
noise. Each test administration is indexed by four perturbation
parameters: the layer \(\ell \in \{0, \dots, 39\}\) at which the noise
is applied, the standard deviation \(\sigma\) of the multiplicative
Gaussian noise (the magnitude of the perturbation, loosely analogous to
the depth of focal tissue damage in patients), the perturbation density
\(\rho \in [0, 1]\) (the fraction of the layer's weights that are
perturbed, loosely analogous to lesion extent), and a random seed \(s\)
governing the realization. For each weight \(w\) selected for
perturbation we set \(w \to w \cdot (1 + \varepsilon)\), with
\(\varepsilon \sim N(0, \sigma^2)\) drawn independently per weight;
\(\rho\) is the fraction of the layer's weights, selected at random
under seed \(s\), to which this is applied, with the remainder left
intact. Each (layer, \(\sigma\), \(\rho\), seed) cell yields a vector of
error proportions across the seven PNT categories. Only the layer axis
is treated as an ordered spatial dimension for cluster correction
(analogous to the way fMRI cluster correction treats the cortical
lattice). Noise magnitude \(\sigma\) and density \(\rho\) index the
depth and extent of the perturbation rather than a position; we
therefore aggregate them within each seed before group-level spatial
inference, and treat the seed itself as the subject in a group analysis.
We use a discovery set and an independent validation set of equal size
(40 seeds each, 80 total).

\textbf{LLM subtraction analysis (two-level structure)}

The LLM-side analysis has a two-level structure within each seed. The
lower level, the \(\sigma \times \rho\) block of contrast values at a
given layer within a given seed, carries the dose-response information
(perturbation depth and extent) and is itself a result, not a
preprocessing step. The upper level, the per-seed layer profile that
feeds the spatial-extent inference step, is obtained from an explicit
linear model of the lower-level cells. The two levels feed two separate
analyses, both reported below.

Lower level (within seed). For each ordered category pair \((A, B)\) and
each seed \(s\), the per-cell contrast is
\(D_s(\ell, \sigma, \rho) = p_s(A \mid \ell, \sigma, \rho) - p_s(B \mid \ell, \sigma, \rho)\).
Within seed, we fit the linear model

\[
D_s(\ell, \sigma, \rho) \;=\; \alpha_s(\ell) \;+\; \beta_{s,\sigma}\,\sigma \;+\; \beta_{s,\rho}\,\rho \;+\; \epsilon,
\]

with layer carried as a categorical regressor (40 intercepts
\(\alpha_s(\ell)\)) and \(\sigma\) and \(\rho\) as continuous
within-seed repeated-measures covariates. The per-(seed, layer) summary
used by the spatial step is the layer-specific fitted value at
cohort-mean \(\sigma\) and \(\rho\),

\[
D_s(\ell) \;=\; \hat\alpha_s(\ell) \;+\; \hat\beta_{s,\sigma}\,\bar\sigma \;+\; \hat\beta_{s,\rho}\,\bar\rho.
\]

This is mathematically equivalent to the simple within-seed average over
the \(\sigma \times \rho\) grid (the cohort means \(\bar\sigma\) and
\(\bar\rho\) recover the average), but the explicit fit makes clear that
\(\sigma\) and \(\rho\) are kept in the model rather than aggregated
away. The retained slopes \(\beta_{s,\sigma}\) and \(\beta_{s,\rho}\)
feed the separate dose-response analysis described in Methods §``Stage
2.5: Dose-response analysis.''

Upper level (across seeds). The seed plays the role of subject.
Group-level inference is computed across discovery seeds on the per-seed
layer profiles: the layer-wise mean \(\bar D(\ell)\) and one-sample
t-statistic against zero, \(T(\ell)\), are computed with NaN-safe
estimators. Spatial-extent inference is enforced by threshold-free
cluster enhancement (TFCE; Smith and Nichols 2009) applied to
\(T(\ell)\) along the layer axis. Because subtraction has a meaningful
sign, we apply TFCE separately to the positive and negative halves of
\(T(\ell)\) and recombine. Surviving layer clusters are extracted by
connected-component labeling at the 90th percentile of the absolute TFCE
field. \textbf{Only the layer axis is used as an ordered spatial
dimension for cluster correction}; \(\sigma\) and \(\rho\) remain dose
parameters and re-enter the analysis in the dose-response step.

\textbf{Human subtraction analysis}

On the human side, we apply the same subtraction logic to the cortex,
parameterized at the level of left-hemisphere Johns Hopkins University
(JHU) atlas regions (Mori et al.~2008). Univariate lesion-symptom
mapping is the established baseline for relating behavioral deficits to
focal damage in chronic aphasia (Bates et al.~2003; Rorden et al.~2007;
Karnath et al.~2018), and atlas-based ROI parcellation rather than
voxel-wise inference is a standard choice when patient sample sizes are
modest and when per-voxel lesion coverage varies sharply across the
brain (Bonilha et al.~2014; Yourganov et al.~2016; Pustina et al.~2018).
The pairwise-contrast logic that motivates the LLM side has direct
precedent on the cortex: Schwartz and colleagues established that two
distinct error classes in the PNT (semantic versus phonemic) load on
dissociable cortical territories when their lesion-symptom maps are
contrasted against one another (Schwartz et al.~2009, 2012; Walker et
al.~2011), an approach extended by Mirman et al.~(2015) to a full
inventory of post-stroke language deficits. We adopt the same contrast
logic at the ROI level.

For each error category \(A\) and each left-hemisphere ROI \(r\), we
compute the Spearman rank correlation \(\rho_{A,r}\) between the
per-patient proportion of category-\(A\) errors and the per-patient
lesion load in \(r\). The Spearman correlation is robust to outliers in
the heavy-tailed distributions of single-patient error proportions, a
known concern in PNT data with high inter-patient variability (Bates et
al.~2003; Rorden et al.~2007). For each ordered category pair \((A, B)\)
and each ROI \(r\), the cortical subtraction value is the difference of
correlations \(d_r = \rho_{A,r} - \rho_{B,r}\): positive \(d_r\) marks
regions where lesions preferentially elevate \(A\)-type errors relative
to \(B\), negative \(d_r\) marks the reverse.

Per-correlation two-sided p-values are obtained from
\texttt{scipy.stats.spearmanr}. Inference on the correlation difference
uses a patient-level percentile bootstrap (1,000 resamples) on \(d_r\),
the standard nonparametric tool when the joint distribution of two
related effect-size estimates is not analytic (Efron and Tibshirani
1993; Sperber and Karnath 2018; DeMarco and Turkeltaub 2018). This is
the cortical analog of the LLM voxel-wise subtraction described above
and is the same calculation rendered in the SNL\_2026\_SLM interactive
viewer (\texttt{simpleSubtraction}). Full atlas, ROI-selection, and
bootstrap details are reported in the \emph{Supplementary Methods}.

\textbf{Out-of-sample replication and Stage 2 permutation validation}

For the LLM side, the seed-as-subject pipeline is fit to the 40-seed
discovery set and tested on the 40-seed held-out validation set: we
recompute \(\bar D(\ell)\) and the layer-axis TFCE on validation seeds
and check whether each discovery-surviving layer cluster preserves its
sign and shape. For the human side, the matched analog is a 50/50
patient split (106 discovery / 107 validation; fixed random seed) of the
213-patient cohort. We recompute per-(error, ROI) Spearman correlations
and per-(pair, ROI) correlation differences on each half, and a (pair,
ROI) test is considered to replicate when its sign matches between
halves and its patient-level bootstrap 95\% CI excludes zero on both
halves. The three pre-registered confirmatory contrasts on both
substrates are Semantic versus Phonemic as the direct test of
dissociation between the two clinically central real-word error classes,
and Phonemic versus Neologism and Semantic versus Neologism as
cross-checks of each cluster against a non-real-word baseline.

PRISM Stage 2 is a layer-order permutation test that licenses the
layer-axis TFCE claim: for each pair, we recompute the group-level
layer-axis TFCE map after independently permuting layer labels within
each seed (which preserves the per-seed marginal distribution of
contrast values but destroys the intrinsic ordering of layers). The
empirical \(p\)-value per (pair, sign-direction) is the proportion of
1,000 random permutations whose maximum \(|\mathrm{TFCE}|\) exceeds the
observed value. Significance under this null indicates that the cluster
mass depends on the intrinsic layer ordering rather than on the marginal
distribution alone. PRISM Stage 3 (confirmatory ROI perturbation against
matched non-significant control clusters) is an LLM-only construct by
design, there is no patient-side analog, and is deferred to a follow-up
study; PRISM as reported in this paper is therefore an LLM method with a
matched human comparison at Stages 1 and 2, rather than a symmetric
two-substrate framework at all three stages.

The full TFCE definition and parameters, the formal bootstrap /
multiple-comparison procedures, and the per-pair region-summary tables
are reported in the \emph{Supplementary Methods}.

\textbf{Stage 2.5: Per-cluster dose-response analysis}

Each surviving layer cluster is the input to a separate dose-response
analysis that re-introduces \(\sigma\) and \(\rho\) as the
perturbation-level analogs of lesion severity and lesion extent. For
each cluster (layers \(L^\star\), pair \((A, B)\), sign), per seed \(s\)
we restrict to layers \(L^\star\) and take the within-cluster mean
contrast at each \((\sigma, \rho)\) cell,

\[
D_s(\sigma, \rho \mid L^\star) \;=\; \mathrm{mean}_{\ell \in L^\star}\!\bigl[ p_s(A \mid \ell, \sigma, \rho) - p_s(B \mid \ell, \sigma, \rho) \bigr].
\]

Across seeds, we compute per-cell group means
\(\bar D(\sigma, \rho \mid L^\star)\). We then fit an OLS dose-response
model on the per-cell means,

\[
\bar D(\sigma, \rho \mid L^\star) \;\sim\; \beta_0 \;+\; \beta_\sigma\,\sigma \;+\; \beta_\rho\,\rho \;+\; \beta_{\sigma\rho}\,\sigma\rho,
\]

and apply a 1,000-iterate cluster bootstrap over seeds for 95\%
confidence intervals on each slope. The conceptual prediction is
directional: \(\sigma\) (perturbation depth) is the analog of how
completely tissue is destroyed within a damaged region in human lesions,
so we expect \(\beta_\sigma\) to be of the same sign as the cluster's
contrast on average (deeper perturbation → larger directional contrast).
\(\rho\) (perturbation extent) is the analog of lesion volume, and its
predicted effect is less constrained: higher density could either deepen
the contrast at the same layers or broaden it across more layers.

We deliberately report dose-response effects as effect-size estimates
and surface shapes rather than as p-values. At 40 discovery seeds × 110
\((\sigma, \rho)\) cells, and because these cells are repeated
measurements within the same seeds rather than independent observations,
formal p-values for small non-zero slopes are far less informative than
the direction, magnitude, uncertainty, replication, and shape of the
estimated surface; the claims we emphasize are therefore the
\emph{direction} of the slope (does the contrast scale as a graded
lesion would?) and the \emph{shape} of the dose-response surface
(monotonic across the full perturbation range, or saturating?). The same
dose-response analysis is repeated on the held-out 40-seed validation
cohort to verify that the recovered surface shapes replicate.

\textbf{Results}

\includegraphics[width=1\textwidth,height=\textheight]{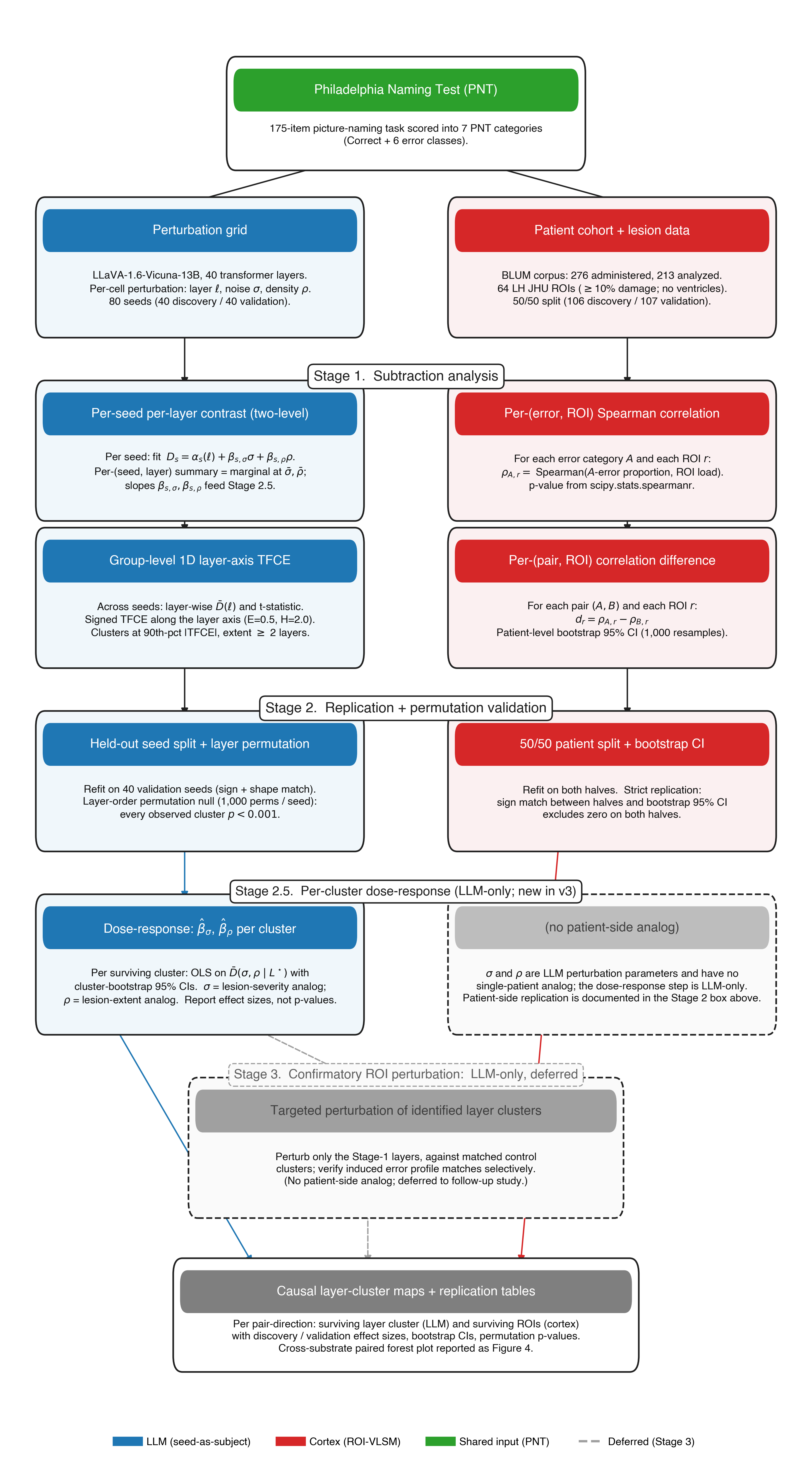}

\textbf{Figure 1.} Schematic of the PRISM pipeline. Top branch (LLM,
seed-as-subject layer-axis TFCE): per perturbation seed \(s\) and each
(layer \(\ell\), noise standard deviation \(\sigma\), perturbation
density \(\rho\)) cell, score the 158-item baseline-correct analysis set
into seven categories; for each ordered pair \((A, B)\) average the
per-cell difference \(p_s(A) - p_s(B)\) across the
\(\sigma \times \rho\) grid within seed to yield a per-seed layer
profile \(D_s(\ell)\); group-level mean, t-statistic, and 1D TFCE along
the layer axis across seeds; replicate on a held-out 40-seed split.
Bottom branch (cortex, ROI-level Spearman correlation-difference VLSM):
per patient score the PNT into seven categories and compute the per-ROI
lesion-load fraction across 64 left-hemisphere JHU atlas regions;
per-(error, ROI) Spearman rank correlation across patients; per-(pair,
ROI) correlation difference \(d_r = \rho_{A,r} - \rho_{B,r}\); replicate
on a 50/50 patient split with patient-level bootstrap. The pipelines are
matched in subject dimension (seeds / patients), spatial dimension
(transformer layers / atlas-parcellated cortex), and thresholding step
(TFCE along an ordered axis / correlation-difference VLSM with bootstrap
CI); the contrast operator differs (within-subject error-proportion
difference / between-subject correlation difference).

We applied PRISM to the three pre-registered confirmatory error
contrasts on both substrates. Semantic versus Phonemic is the direct
test, pitting the two clinically central real-word error classes against
each other; Phonemic versus Neologism and Semantic versus Neologism each
cross-check one of the two real-word classes against the non-real-word
baseline that Neologism provides. The same three contrasts were run on
the human cortex (213 patients, 50/50 split into 106 discovery and 107
validation) and on the LLM (80 perturbation seeds, 40 discovery and 40
validation), parameterized at the level appropriate to each substrate:
layer-axis TFCE over a seed-as-subject group analysis on the LLM, and
ROI-level univariate Spearman correlation-difference VLSM on the 64
left-hemisphere JHU atlas regions for the cortex. Before reporting the
subtraction analyses, we first verify that every error category carries
a non-trivial univariate signal on both substrates; full per-pair
subtraction results follow in the subsequent subsections (with all 15
pairwise contrasts in Supplementary Figures S2 and S3, and full
inferential detail for the univariate analyses in Supplementary §S7 and
Tables S5--S6).

\textbf{Per-category univariate signal on both substrates}

Subtraction can mask category-specific signal that \emph{is} present at
the level of the marginals, two error categories with similar spatial
loading can dissociate weakly even when each is itself spatially
structured. We therefore first report the per-category univariate maps
on both substrates before any pairwise subtraction (Figure 2). On the
LLM, the per-layer mean error-category proportion across the 40
discovery seeds (with \(\sigma\) and \(\rho\) averaged within seed)
shows a category-specific layer profile for every one of the six
non-Correct PNT categories: Phonemic peaks at deep-middle layers
(\textasciitilde25--30), Semantic rises gradually through the network
and remains elevated at the deepest layers, Neologism peaks in
mid-network (\textasciitilde15--20), Mixed accumulates gradually,
Unrelated peaks at mid-network and falls off, and NoResponse carries a
high baseline that declines across the network. On the cortex, the
per-ROI Spearman rank correlation between each error-category proportion
and the per-patient JHU ROI lesion load, without any age, sex, or other
covariate adjustment, to match the no-covariates LLM-side analysis,
yields category-specific spatial maps for every category as well;
Phonemic and NoResponse show the strongest perisylvian loading, with
Semantic, Neologism, Mixed, and Unrelated each lighting up smaller but
anatomically interpretable territories. The point of this subsection is
not to interpret each map in detail (those interpretations and
per-category top-ROI tables are in Supplementary §S7 and Tables S5--S6)
but to establish that signal is present per-category on both substrates
\emph{before} any subtraction. The subtraction results that follow
therefore reflect \emph{differential} loading across paired categories,
not the presence or absence of signal for either category individually.

\includegraphics[width=1\textwidth,height=\textheight]{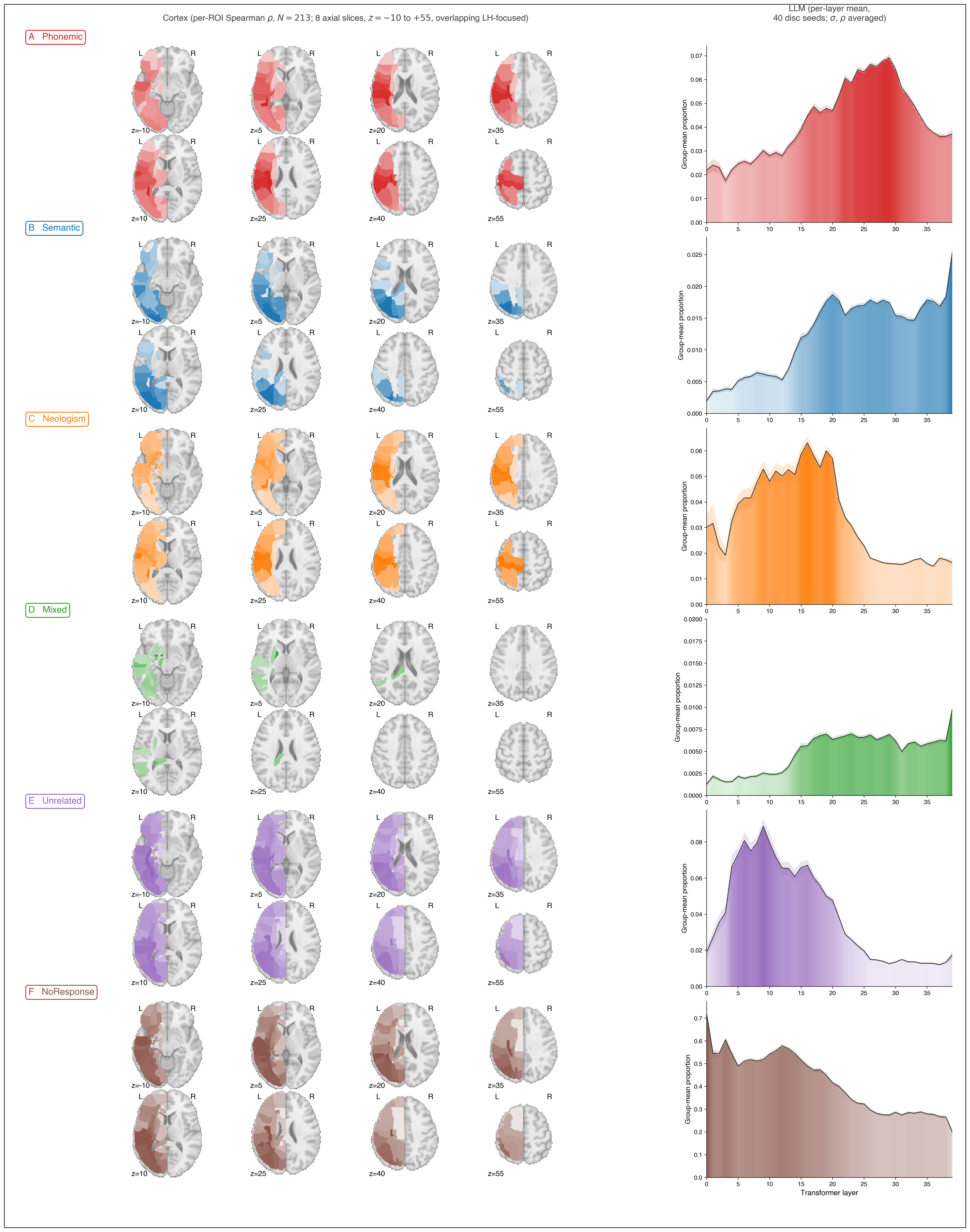}

\textbf{Figure 2.} Per-category univariate signal on both substrates,
computed without covariate adjustment to match the no-covariates LLM
analysis. Layout is 6 rows × 2 columns, one row per non-Correct PNT
error category. Left column (cortex): per-ROI Spearman rank correlation
between the per-patient PNT error-type proportion and the per-patient
JHU ROI lesion load on the \(N = 213\) analyzed patients, rendered on
three axial slices of the MNI152 T1 template at \(z = -2, +18, +38\).
Right column (LLM): per-layer group-mean PNT proportion across the 40
discovery seeds with \(\sigma\) and \(\rho\) averaged within seed, with
95\% CI shading. \textbf{A.} Phonemic. \textbf{B.} Semantic. \textbf{C.}
Neologism. \textbf{D.} Mixed. \textbf{E.} Unrelated. \textbf{F.}
NoResponse. Each error category carries category-specific signal on both
substrates; the subtraction results that follow (Figures 3 and 4) report
the \emph{differential} loading across paired categories rather than the
presence or absence of signal at the marginal.

\textbf{Human cortex: a robust frontal-perisylvian Phonemic
\textgreater{} Semantic cluster; the Semantic \textgreater{} Phonemic
direction is a consistently signed but non-significant trend}

On the patient cohort, the primary Semantic-versus-Phonemic contrast
yields an unambiguous Phonemic \textgreater{} Semantic ROI cluster
centered on the left postcentral gyrus
(\(\Delta\rho_{\mathrm{disc}} = -0.275\), bootstrap 95\% CI
\([-0.568, -0.012]\) on discovery,
\(\Delta\rho_{\mathrm{val}} = -0.301\), \([-0.512, -0.058]\) on
validation), precentral gyrus (disc \(-0.234\), \([-0.521, +0.045]\);
val \(-0.299\), \([-0.531, -0.064]\)), and SLF (disc and val both
significantly negative). The Phonemic \textgreater{} Semantic direction
at PoCG\_L and PrCG\_L is the only direction--ROI combination in the
primary contrast that survives the strictest replication criterion,
sign-matched across halves with both discovery and validation bootstrap
95\% CIs excluding zero. The Semantic \textgreater{} Phonemic direction
has consistently signed point estimates on discovery in posterior
temporo-occipital cortex (LG\_L, Cu\_L, MOG\_L, SOG\_L, IOG\_L) but the
validation point estimates flip sign or shrink to near zero, and no
Semantic \textgreater{} Phonemic ROI's CI excludes zero on both halves.
The honest characterization is therefore that the Phonemic
\textgreater{} Semantic direction is a robust replicating cluster on the
cortex and the Semantic \textgreater{} Phonemic direction is a
consistently signed but non-significant trend (Figure 3). The
Phonemic-versus-Neologism and Semantic-versus-Neologism cross-checks
recover the same dissociation: Neologism \textgreater{} Semantic loads
on the same frontal-perisylvian territory as Phonemic \textgreater{}
Semantic (PoCG, PrCG, SLF all replicating), and the semantic-favoring
direction in the cross-checks is again a sub-significant trend. Full
per-pair ROI maps and statistics for all 15 pairwise contrasts are in
Supplementary Figure S3 and Tables S2--S3.

\includegraphics[width=1\textwidth,height=\textheight]{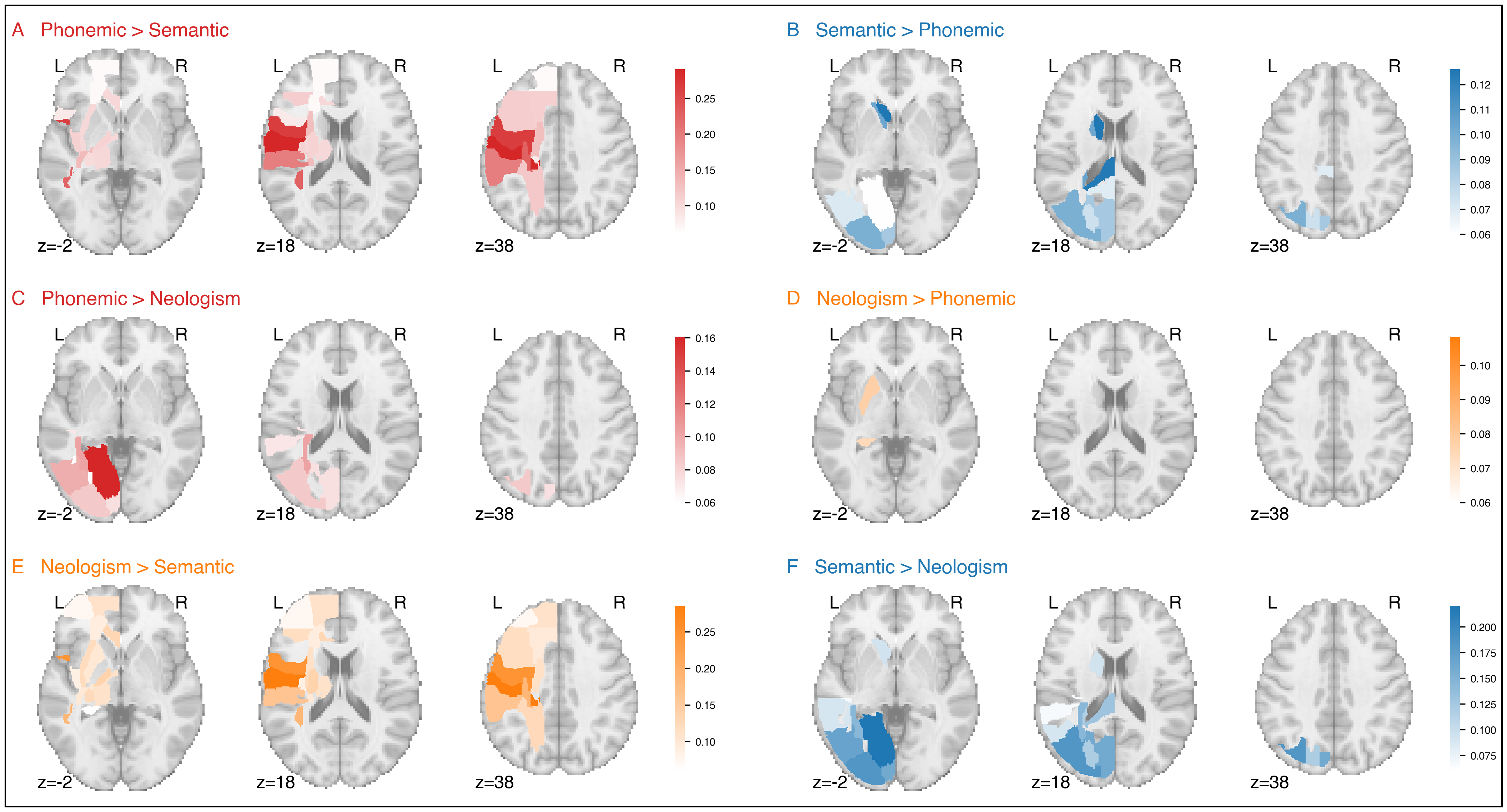}

\textbf{Figure 3.} Human LSM cortical subtraction maps for the three
confirmatory contrasts on the patient cohort. Each left-hemisphere JHU
atlas region is colored by its correlation difference
\(\Delta\rho_r = \rho(A, \mathrm{lesion}_r) - \rho(B, \mathrm{lesion}_r)\),
where \(\rho\) is the Spearman rank correlation between the per-patient
error-type proportion and the per-patient lesion load in the ROI,
rendered on three axial slices of the MNI152 T1 template
(\(z = -2, 18, 38\)). Color matches the category favored by each
direction (red = Phonemic-favoring, blue = Semantic-favoring, orange =
Neologism-favoring). The six direction-split panels here pair into the
three confirmatory contrasts of Figure 4 (panels A+B → Figure 4A; panels
C+D → Figure 4B; panels E+F → Figure 4C).

\textbf{A.} Phonemic \textgreater{} Semantic (primary contrast).
\textbf{B.} Semantic \textgreater{} Phonemic (primary contrast).
\textbf{C.} Phonemic \textgreater{} Neologism (cross-check). \textbf{D.}
Neologism \textgreater{} Phonemic (cross-check). \textbf{E.} Neologism
\textgreater{} Semantic (cross-check). \textbf{F.} Semantic
\textgreater{} Neologism (cross-check).

The three phonemic- and neologism-favoring panels (A, C, E) all light up
the same left frontal-perisylvian territory, postcentral gyrus,
precentral gyrus, supramarginal gyrus, and superior longitudinal
fasciculus. The three semantic-favoring panels (B, D, F) are markedly
weaker and more diffusely distributed, an asymmetry that parallels the
LLM-side result below (Figure 4).

\textbf{LLM layers: a robust deep Phonemic \textgreater{} Semantic
cluster; the Semantic \textgreater{} Phonemic direction does not
survive}

The three confirmatory contrasts on the LLM side, run under the
seed-as-subject layer-axis TFCE pipeline, recover the same asymmetric
pattern (Figure 4). The primary Semantic-versus-Phonemic contrast yields
a single robust Phonemic \textgreater{} Semantic layer cluster spanning
layers 22--31 on discovery (TFCE peak at layer 22, with the cluster's
largest \(\bar D \approx -0.05\) at layer 29; Stage-2 layer-permutation
\(p < 0.001\)) and 23--33 on validation, with no surviving Semantic
\textgreater{} Phonemic cluster in either half (Figure 4A). The
Phonemic-versus-Neologism cross-check yields a Phonemic \textgreater{}
Neologism cluster at layers 24--31 (Figure 4B; peak
\(\bar D \approx +0.05\)) coincident in depth with the primary-contrast
Phonemic \textgreater{} Semantic cluster, and only a
consistent-sign-but-non-significant Neologism \textgreater{} Phonemic
trend in earlier layers. The Semantic-versus-Neologism cross-check
yields two Neologism \textgreater{} Semantic clusters at layers 16--17
and 19--20 (Figure 4C) and essentially no Semantic \textgreater{}
Neologism layer cluster. Across all three contrasts, the surviving layer
clusters all sit in the Phonemic- and Neologism-favoring directions; the
Semantic-favoring direction is a consistently signed but sub-threshold
trend on the LLM side, matching the human-side asymmetry. Stage 2
layer-order permutation validation: every observed cluster across the 15
pairwise contrasts cleared \(p < 0.001\) against an
independently-permuted-layer null, confirming that the cluster mass
depends on the intrinsic layer ordering and not on the marginal
distribution of contrast values.

\includegraphics[width=1\textwidth,height=\textheight]{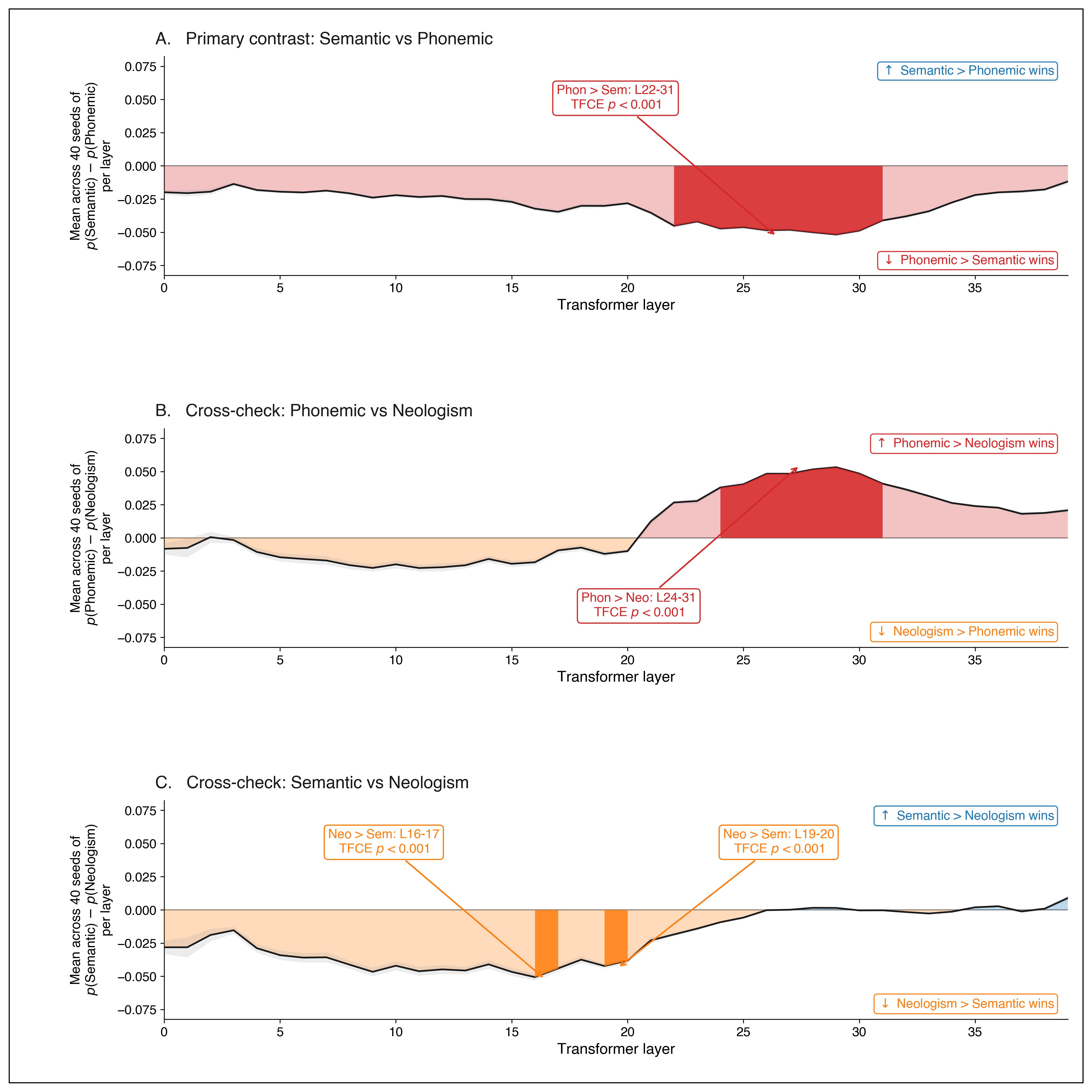}

\textbf{Figure 4.} PRISM layer maps for the three confirmatory contrasts
on the LLM, organized as three pair-level panels that pair into the
direction-split panels of Figure 3 (Figure 4A ↔ Figure 3 A+B; Figure 4B
↔ Figure 3 C+D; Figure 4C ↔ Figure 3 E+F). Each panel plots the
group-level layer-axis profile
\(\mathrm{mean}_{\sigma, \rho}\,[p(\mathrm{cat}_1) - p(\mathrm{cat}_2)]\)
across the 40 discovery seeds; the area between the curve and zero is
filled in the category color of whichever side is favored at each layer
(above zero = the first-listed category wins; below zero = the
second-listed wins). TFCE-surviving cluster ranges are highlighted with
a saturated solid fill and a labeled callout in the opposite (empty)
half-plane, giving the cluster's layer range and the Stage-2
layer-permutation \(p\)-value. Directional labels inside each panel
indicate which category is favored by each sign of the y-axis.

\textbf{A.} Primary contrast, Semantic vs Phonemic. The curve is below
zero throughout the network, with a TFCE-surviving Phonemic
\textgreater{} Semantic cluster at layers 22--31. \textbf{B.}
Cross-check, Phonemic vs Neologism. The curve crosses zero around layer
20; a TFCE-surviving Phonemic \textgreater{} Neologism cluster sits at
layers 24--31. \textbf{C.} Cross-check, Semantic vs Neologism. The curve
is below zero for most of the network, with two TFCE-surviving Neologism
\textgreater{} Semantic clusters at layers 16--17 and 19--20.

The semantic-favoring direction does not survive cluster correction on
either the discovery or the validation half, so a natural skeptical
reading is that the Semantic \textgreater{} Phonemic effect is real but
underpowered. Three observations argue against that reading and for the
asymmetry as a feature of the data rather than a sampling artifact.
First, the Sem-favoring point estimates do not grow from discovery to
validation as a power-limited effect would: \(\bar D\) in the
Sem-favoring direction is indistinguishable from zero on both halves,
with no discernible directional drift across the seed split (Figure 5;
Supplementary Table S1). Second, the same asymmetry replicates on the
cortical analysis, where \(N = 213\) patients dwarfs the 40-seed LLM
discovery cohort; if the LLM null were purely a power problem we would
expect the larger-\(N\) cortical analysis to surface the Sem-favoring
effect clearly, and it does not (no Sem-favoring ROI clears the strict
replication criterion on the 50/50 patient split; Tables S2, S3). Third,
the dose-response analysis (below) finds no positive scaling of the
contrast magnitude with \(\sigma\) or \(\rho\) in the Sem-favoring
direction at any surviving cluster's layers, ruling out the ``signal is
present but obscured by noise'' version of the power explanation. We
nevertheless flag in Future Directions that a higher-seed-count
replication is planned as a direct power test; the planned design scales
the discovery and validation seed counts beyond 40 each while holding
the analysis pipeline fixed.

\textbf{Out-of-sample replication on both substrates}

LLM side: the discovery-set layer clusters reproduce on the 40-seed
held-out validation set. The primary Semantic-versus-Phonemic Phonemic
\textgreater{} Semantic cluster shifts only marginally in extent
(discovery: layers 22--31; validation: layers 23--33) and not in sign or
location; the Phonemic \textgreater{} Neologism cluster on validation
lands at layers 24--33 (discovery: 24--31), again preserving sign and
location. The Semantic \textgreater{} Phonemic and Semantic
\textgreater{} Neologism directions remain sub-threshold on validation,
as they did on discovery. Stage 2 layer-permutation: all 15 observed
pair-level clusters cleared the empirical \(p < 0.001\) threshold
against an independently permuted-layer null, demonstrating that the
cluster mass is licensed by the intrinsic ordering of layers.

Human side: of the \(15 \times 64 = 960\) (pair, ROI) tests in the 50/50
patient split, 714 (74\%) had matching sign between discovery and
validation halves, and 60 (6.3\%) survived the strict replication
criterion of sign match + bootstrap CI excluding zero on both halves. In
the primary contrast, the strictly replicating ROIs are exclusively in
the Phonemic \textgreater{} Semantic direction (PoCG\_L: disc
\(\Delta\rho = -0.275\) {[}-0.568, -0.012{]}, val \(-0.301\) {[}-0.512,
-0.058{]}; PrCG\_L: disc \(-0.234\) {[}-0.521, +0.045{]}, val \(-0.299\)
{[}-0.531, -0.064{]}); no Semantic \textgreater{} Phonemic ROI in the
primary contrast replicates by this criterion. The same asymmetry is
mirrored in both Neologism cross-checks. The cross-substrate headline is
consistent: the phonemic-favoring direction is robust and replicates on
both substrates; the semantic-favoring direction is a consistently
signed but non-significant trend on both substrates. Full per-(pair,
ROI) replication statistics are in Supplementary Table S2.

\includegraphics[width=1\textwidth,height=\textheight]{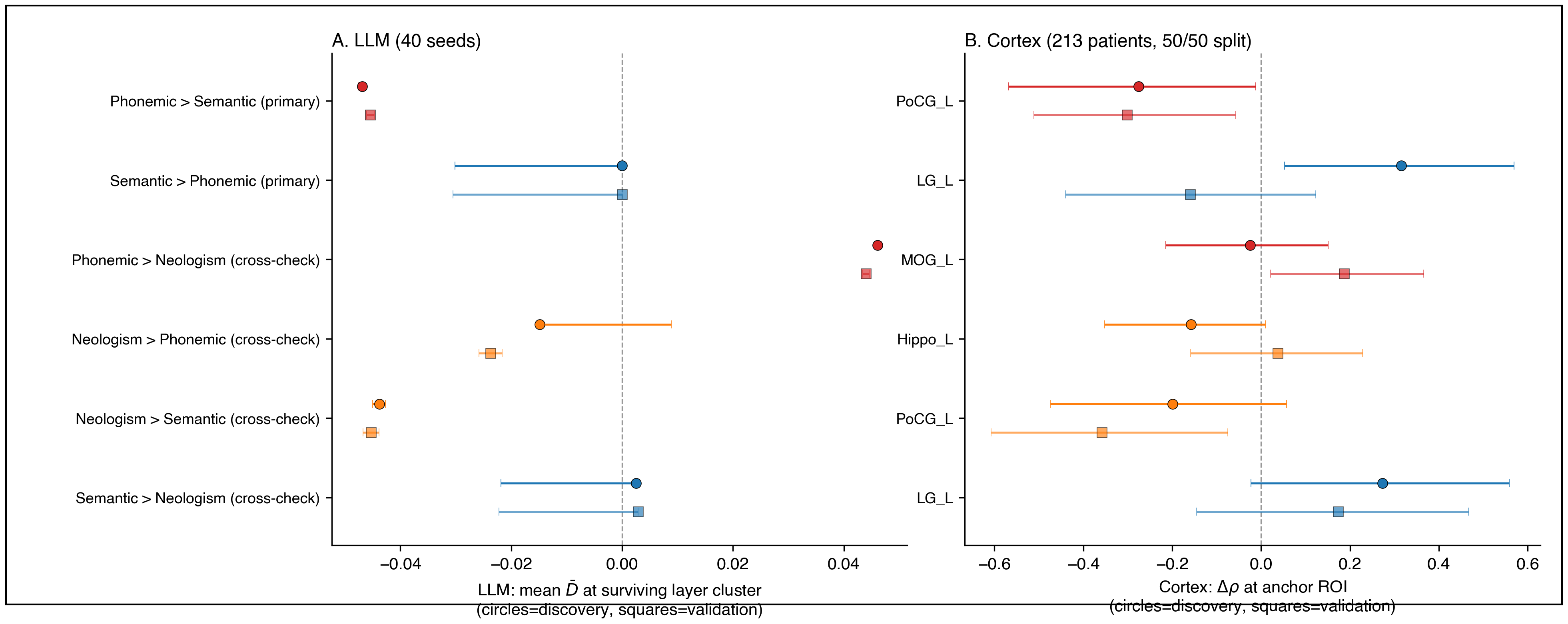}

\textbf{Figure 5.} Cross-substrate forest plot. Per pair-direction
discovery and validation \(\bar D\) at the surviving layer cluster on
the LLM side, and discovery and validation \(\Delta\rho\) for the
strictest-replicating ROIs on the human side, each with 95\% bootstrap
confidence intervals. The plot compresses the asymmetry into a single
visual: the phonemic-favoring effects (Phonemic \textgreater{} Semantic
primary; Phonemic \textgreater{} Neologism cross-check) cluster on the
negative-\(\Delta\) side on both substrates and have CIs excluding zero
on both halves, while the semantic-favoring effects sit at or near zero
on both substrates.

\textbf{Dose-response structure of the recovered layer clusters}

Each surviving phonemic-favoring layer cluster on the LLM side carries a
dose-response signature with respect to perturbation depth (\(\sigma\))
and perturbation extent (\(\rho\)), the LLM analogs of lesion severity
and lesion extent (Figure 6; full per-pair slopes for all 15 pairwise
contrasts on both halves are in Supplementary Table S4). For the primary
Semantic-vs-Phonemic cluster (layers 22--31), the discovery-seed OLS
dose-response slopes are \(\hat\beta_\sigma = -0.182\) (95\%
cluster-bootstrap CI \([-0.185, -0.179]\)) and
\(\hat\beta_\rho = -0.462\) (\([-0.474, -0.451]\)), both with the same
sign as the cluster's contrast: deeper and more extensive perturbations
produce a stronger Phonemic \textgreater{} Semantic dissociation on
average. The same slopes on validation seeds are essentially identical
(\(\hat\beta_\sigma = -0.176\), \(\hat\beta_\rho = -0.434\)). The
Phonemic-vs-Neologism cross-check shows the same scaling pattern at
coincident layers (24--31); the Semantic-vs-Neologism Neologism
\textgreater{} Semantic effect at the early mid-network layers (19--20)
scales in the same (negative) direction as its contrast, so the
Neologism \textgreater{} Semantic dissociation likewise strengthens with
both \(\sigma\) and \(\rho\).

The dose-response surface in Figure 6 is informative beyond the linear
slope. Across all three clusters, the per-cell mean \(|D|\) does not
increase monotonically with \(\sigma\) and \(\rho\) across the full
grid. It peaks along a diagonal ridge on which perturbation depth and
perturbation extent trade off inversely, and falls away on both sides of
that ridge: for the primary cluster the global maximum sits at
\(\sigma = 1.5\), \(\rho = 0.7\) (\(|D| = 0.22\)), with comparable
maxima at \(\sigma = 1.8\), \(\rho = 0.4\) and \(\sigma = 2.0\),
\(\rho = 0.3\), while both the low-dose corner (\(\sigma = 1.1\),
\(\rho = 0.1\)) and the high-dose corner (\(\sigma = 2.0\),
\(\rho = 1.0\)) fall to near zero. What governs the contrast is
therefore the total amount of perturbation delivered rather than either
parameter alone. We interpret this as a category-differentiation regime:
moderate total perturbation maximally separates error categories from
each other, while severe perturbation degrades that differentiation as
all error categories rise together. The directional slopes
(\(\hat\beta_\sigma\) and \(\hat\beta_\rho\), each matching its
cluster's contrast sign) are coarse directional summaries of the average
dose dependence across the grid; the bilinear-plus-interaction OLS
explains only a modest share of each surface
(\(R^2 \approx 0.18\)--\(0.26\)), so we treat \(\hat\beta_\sigma\) and
\(\hat\beta_\rho\) as directional descriptors of the observed
\(\sigma \times \rho\) surfaces rather than as a global model of them.
The saturating shape captures the limit of the perturbation-as-lesion
analogy at very high perturbation levels. Both properties replicate on
the held-out 40-seed validation cohort.

\includegraphics[width=1\textwidth,height=\textheight]{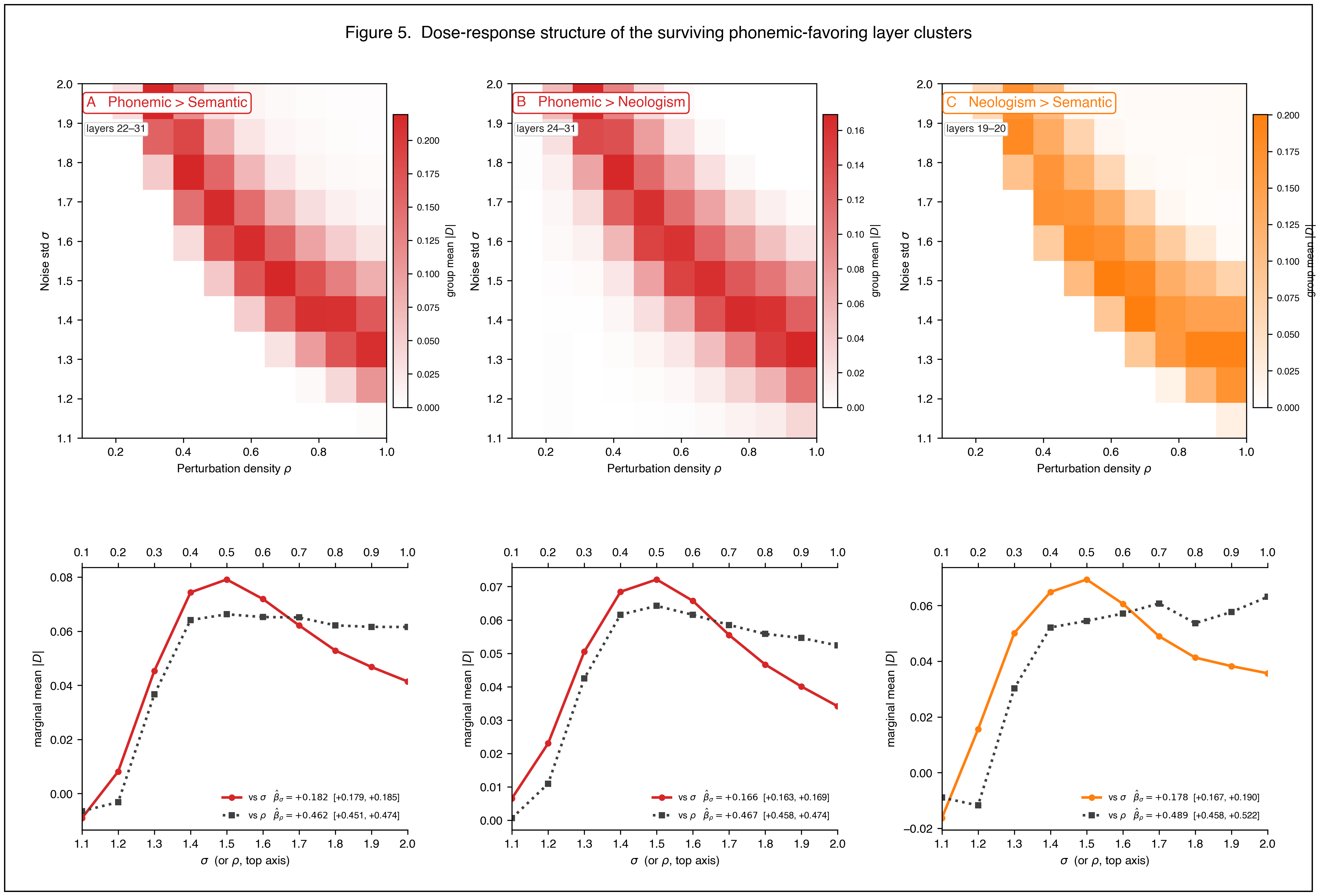}

\textbf{Figure 6.} Per-cluster dose-response surfaces for the three
surviving phonemic-favoring layer clusters on the LLM side, computed
from the 40 discovery seeds. Top row: heatmaps of group-mean \(|D|\) at
the cluster's layers across the \(\sigma \times \rho\) perturbation
grid. Bottom row: marginal mean \(|D|\) as a function of \(\sigma\)
(solid, panel anchor color) and as a function of \(\rho\) (dotted,
grey), with cluster-bootstrap 95\% CIs on the OLS slopes
\(\hat\beta_\sigma\) and \(\hat\beta_\rho\). \textbf{A.} Phonemic
\textgreater{} Semantic (primary; Semantic\_minus\_Phonemic, negative
direction, layers 22--31). \textbf{B.} Phonemic \textgreater{} Neologism
(cross-check; Phonemic\_minus\_Neologism, positive direction, layers
24--31). \textbf{C.} Neologism \textgreater{} Semantic (cross-check;
Semantic\_minus\_Neologism, negative direction, layers 19--20). Across
all three clusters, each cluster's directional contrast scales in the
direction of its own sign, deepening with both \(\sigma\) (perturbation
depth, analog of lesion severity) and \(\rho\) (perturbation extent,
analog of lesion volume) over the lower part of the grid, then declining
once total perturbation passes a moderate level, consistent with a
graded-lesion regime that gives way to undifferentiated noise at extreme
perturbation.

\textbf{Discussion}

\textbf{Summary of principal findings}

PRISM demonstrates that the subtractive approach can be applied both to
a 13-billion-parameter vision-language transformer and to a cohort of
213 chronic-aphasia patients administered the same task, with the
parallel holding at the level of subject (seeds for the LLM, patients
for the cortex), spatial dimension (transformer layers for the LLM,
atlas-parcellated cortex for the patients), and TFCE-style thresholding
along an ordered axis. The contrast operator is not parallel, the LLM
side is a within-subject difference in error proportions averaged across
seeds, the human side is a between-subject correlation difference, but
the spatial-inferential machinery is. On the LLM side, the primary
Semantic-versus-Phonemic contrast under seed-as-subject layer-axis TFCE
yields a robust Phonemic \textgreater{} Semantic layer cluster at layers
22--31 (discovery, validation 23--33), which is corroborated by a
coincident Phonemic \textgreater{} Neologism cluster at layers 24--31
against the non-real-word baseline. The Semantic \textgreater{} Phonemic
direction does not survive on the LLM in any of the three confirmatory
contrasts. PRISM Stage 2 layer-order permutation testing places every
observed cluster at empirical \(p < 0.001\) against a null in which
layer order is destroyed within each seed. On the cortex, the matched
ROI-level correlation-difference VLSM with 50/50 patient split recovers
the same asymmetry: PoCG\_L and PrCG\_L survive the strictest
replication criterion in the Phonemic \textgreater{} Semantic direction
on both halves; the Semantic \textgreater{} Phonemic direction has
consistently signed point estimates that do not survive replication. The
honest cross-substrate headline is therefore that the phonemic-favoring
direction is robust and replicates on both substrates, and the
semantic-favoring direction is a consistently signed but non-significant
trend on both substrates, an asymmetry that is itself a replicable
property of the data rather than a failure of the method.

\textbf{Convergence with the human LSM maps}

The Semantic-versus-Phonemic, Phonemic-versus-Neologism, and
Semantic-versus-Neologism contrasts each recover a phonemic-favoring
direction that survives both substrates' replication standards
(layer-axis TFCE with seed-as-subject group analysis on the LLM;
ROI-level Spearman correlation-difference VLSM with 50/50 patient split
on the cortex). On the cortex, the recovered phonemic-favoring territory
recapitulates an effect that has been established for decades in
lesion-symptom mapping of post-stroke aphasia (Bates et al.~2003; Mirman
et al.~2015; Schwartz et al.~2006, 2009, 2012; Dell et al.~1997; Walker
et al.~2011; Lambon Ralph et al.~2017; Patterson et al.~2007). On the
LLM, the same effect is, to our knowledge, new at the layer level: a
transformer hierarchy trained only on next-token prediction reproduces
the phonemic-favoring axis when systematically perturbed, without any
supervision on aphasia-symptomatology constructs. The semantic-favoring
direction is a consistently signed but non-significant trend on both
substrates, neither the LLM layer-axis TFCE (which finds no Semantic
\textgreater{} Phonemic cluster on either discovery or validation seeds)
nor the human-side ROI VLSM (which finds no Semantic \textgreater{}
Phonemic ROI that survives the strict replication criterion in the
primary contrast) supports calling the semantic side a ``recovered
cluster'' in the same sense as the phonemic side. The asymmetry itself
is the replicable finding. The cross-substrate alignment is visible
directly by reading the cortical subtraction map (Figure 3) against the
layer subtraction map (Figure 4): both substrates show a robust
phonemic-favoring map (left perisylvian on cortex; layers 22--31 in the
LLM) and a sub-threshold semantic-favoring trend.

The point of running both pipelines within the same paper is not to
interpret the cross-substrate map alignment in detail, but to
demonstrate that the same conceptual machinery yields the same
asymmetric result on both sides at the inference unit appropriate to
each substrate; the population-level correspondence reported in BLUM
(Fridriksson et al.~2026) is thereby demonstrated inside the present
study rather than asserted by analogy. The cross-substrate result
strengthens the alignment claim made by the broader corpus of LLM--brain
alignment work (Schrimpf et al.~2021; Caucheteux and King 2022;
Goldstein et al.~2022, 2025; Antonello et al.~2024; Tuckute et al.~2024;
Kumar et al.~2024; Mischler et al.~2025; Gao et al.~2025; Mahowald et
al.~2024) by adding behavioral-failure-mode alignment to the existing
repertoire. Falsifiability is also stronger here than in BLUM, because
both sides have now passed through out-of-sample replication on
independent held-out splits and the LLM side has additionally passed
through Stage-2 layer-permutation validation.

\textbf{Importing lesion-symptom mapping into LLM interpretability}

The dominant interpretability methods (probing classifiers, causal
mediation, activation patching, circuit discovery, sparse autoencoders;
Belinkov 2022; Meng et al.~2022; Geiger et al.~2021; Heimersheim and
Nanda 2024; Conmy et al.~2023; Wang et al.~2023; Cunningham et al.~2023;
Templeton et al.~2024) evaluate claims against benchmarks or internal
reconstruction quality rather than against an externally validated,
behaviorally defined map of what model components are causally necessary
for. PRISM supplies that missing criterion by importing the
cluster-based subtraction pipeline of functional neuroimaging (Petersen
et al.~1988; Friston et al.~1995; Smith and Nichols 2009) and the
lesion-symptom-mapping logic that grounds it in causal necessity (Bates
et al.~2003; Mirman et al.~2015), and by applying the same logic to a
human patient cohort within this study. On the LLM side, a
layer-targeted parameter perturbation is an experimental manipulation,
the error-category proportion is the outcome, and the inference is that
the manipulated layer cluster is necessary in the forward pass for
avoiding that error class, a counterfactual structure that becomes a
strict causal-mechanism claim only after the planned PRISM Stage 3
confirmatory ROI intervention (Future Directions). The human side is, as
in all lesion-symptom mapping, quasi-causal: lesions are observed rather
than assigned, so the cortical maps support causal-necessity claims only
to the degree the LSM literature already licenses. A clinician or
aphasia researcher can score a positive PRISM result without inspecting
any model internals; a negative result is falsifiable in the same sense
a failed lesion-symptom prediction is. Both properties were absent from
the dominant component-level tradition.

\textbf{Dose-response evidence that the perturbation behaves as a graded
lesion}

The within-seed \(\sigma \times \rho\) block at each cluster supplies a
second piece of cross-substrate evidence that goes beyond cluster
recovery. \(\sigma\), the standard deviation of the multiplicative
Gaussian noise applied to the targeted layer's weights, is the LLM
analog of how completely tissue is destroyed within a damaged region in
human lesions; \(\rho\), the fraction of that layer's weights perturbed,
is the analog of lesion volume (the spatial extent of damage). The
aphasia literature has worked on the behavioral signature of these two
lesion properties for decades, and the directional prediction is clear:
a deeper lesion produces a worse functional deficit, and a more
extensive lesion produces a wider deficit. On the LLM side, every
surviving phonemic-favoring layer cluster scales its directional
contrast in the direction of that cluster's own contrast (deepening)
with both \(\sigma\) and \(\rho\) across the full perturbation grid;
this scaling replicates on the held-out 40-seed validation cohort. We
read this as consistent with the perturbation acting as a graded lesion
analog rather than as undifferentiated noise. We are cautious not to
overstate it: monotone sensitivity to perturbation magnitude is not by
itself diagnostic of a lesion-like mechanism, since many non-specific
effects also scale with noise. The more distinctive observation is the
shape of the surface, a graded rise followed by saturation (below),
rather than the bare existence of a slope.

This dose-response result also addresses, from a different angle than
Stage 2 permutation, the residual-connection concern raised in the
Introduction (Elhage et al.~2021). The residual-connection worry is that
a layer-targeted perturbation may not isolate that layer's computation
as cleanly as a focal cortical lesion isolates a cortical region. Stage
2 layer-order permutation answers one form of the worry: the recovered
cluster mass depends on the \emph{ordering} of layers and not on the
marginal distribution of contrast values. The dose-response scaling
answers a separate form of the worry: the recovered cluster's contrast
is \emph{graded} in the dose parameter, not flat, which is what we would
expect from a functioning lesion analog but not from a non-specific
perturbation. The dose-response surface saturates at very high
\(\sigma\) and \(\rho\) (Figure 6), marking the limit of the analogy: at
extreme perturbation the LLM produces all error categories together and
category-differentiating signal degrades, the LLM analog of a lesion so
large it no longer maps onto a specific functional deficit.

\textbf{What layer-cluster claims license}

Because PRISM identifies layer clusters by their behavioral consequence
rather than by their internal computation, its claims are coarse-grained
by design and complement rather than compete with the component-level
interpretability tradition. The implicated layer ranges are necessary in
the forward pass under perturbation, not computationally sufficient as
standalone units; residual connections couple every layer's output to
all preceding layers (Elhage et al.~2021), so the cluster identifies a
necessary path rather than a sufficient module. A natural integration is
to apply sparse-autoencoder feature decomposition or activation patching
to the layers PRISM identifies as causally necessary for a given error
category, transforming the search for fine-grained interpretable
features from an exhaustive sweep into a focused search within
behaviorally validated regions. The same logic applies in reverse: a
component-level claim that a particular layer range carries semantic
information is testable under PRISM by checking whether targeted
perturbation of those layers selectively elevates semantic errors.

The behavioral grounding also licenses the patient-specific digital
twins that motivated the broader research program initiated by BLUM.
Within the present architecture, PRISM identifies the perturbation
conditions whose error profiles most closely match an individual
patient's profile, supplying a layer-cluster-level computational
surrogate on which trial-design simulations and individualized
rehabilitation strategies can be screened in silico. The artificial
setting's experimental controllability, noted in the Introduction, is
what makes it useful as a complement to lesion-symptom mapping in human
patients rather than merely an analogue of it.

\textbf{Limitations}

The present analysis is confined to one architecture, one behavioral
task, and Stages 1 and 2 of the three-stage PRISM framework. Whether the
phonemic-favoring-versus-semantic-favoring dissociation replicates in
pure-text language models, in encoder-decoder architectures, at
different parameter scales, in additional clinical tasks (the WAB-R,
sentence completion, story comprehension), and across additional aphasia
subtypes (Broca's, Wernicke's, conduction, anomic, mixed transcortical,
global) are open empirical questions that the same pipeline supports
without modification but that we do not answer here. Stage 3
(confirmatory ROI perturbation against matched non-significant control
clusters) is by construction an LLM-only step, there is no patient-side
analog, and is deferred to a follow-up study; PRISM as reported in this
paper is therefore an LLM method with a matched human comparison at
Stages 1 and 2, rather than a symmetric two-substrate framework at all
three stages. Residual connections couple every transformer layer's
output to all preceding layers, so the implicated layer clusters are
necessary within the forward pass under perturbation rather than
computationally sufficient as standalone units. Finally, the
seven-category clinical taxonomy collapses sub-categories within
Phonemic and Semantic that would themselves be informative; sub-category
contrasts are a defined extension that the present pipeline supports but
that we do not pursue here.

The two perturbation parameters that index the LLM analysis also have
plausible (though loose) clinical analogs in the patient population. The
Gaussian-noise standard deviation \(\sigma\), which controls how heavily
a layer's weights are perturbed multiplicatively, is analogous to the
depth of focal tissue damage or the completeness of a fiber-tract
disconnection on a per-patient basis. The perturbation density \(\rho\),
which controls the fraction of the layer's weights actually perturbed,
is analogous to the spatial extent of the lesion, that is, the fraction
of an anatomically defined region rendered nonfunctional. We frame these
as analogies rather than equivalences; their utility is that the same
coordinates organize the LLM-side variation and (within limits) the
lesion-load variation across patients, supporting future work that
explicitly co-parametrizes severity and extent on both substrates.

\textbf{Future directions}

The immediate completing extension is PRISM Stage 3, confirmatory ROI
perturbation of the identified phonemic-favoring layer cluster against
matched non-significant control clusters, on the LLM side only. Beyond
this, cross-architecture replication tests whether the phonemic-favoring
layer organization is an emergent property of next-token prediction over
language-rich data or a contingent feature of LLaVA-1.6-Vicuna-13B's
training trajectory; cross-task replication tests whether the same
organization carries across picture-naming, sentence-completion, and
free generation. Methodologically, finer error-category subdivision and
integration with sparse-autoencoder decomposition within the
PRISM-identified layer ranges (Cunningham et al.~2023; Templeton et
al.~2024) extend the framework's inferential reach without changing its
conceptual structure. A direct next step on the inferential side is to
resolve the semantic-favoring trend either by enlarging the patient
cohort, by reducing the seven-category PNT taxonomy to a binary
Semantic-vs-Phonemic outcome via re-scoring, or by switching to a
multivariate VLSM/LLM analysis that pools strength across small
per-region effects.

\textbf{Conclusions}

PRISM imports the cluster-based subtraction pipeline of human cognitive
neuroscience into transformer interpretability and applies it, under
matched subject / spatial / TFCE machinery, to a perturbed
13-billion-parameter vision-language transformer and to a cohort of 213
chronic-aphasia patients administered the same task. The
Semantic-versus-Phonemic contrast and its two Neologism cross-checks
each recover a robust phonemic-favoring map on both substrates, a
frontal-perisylvian Phonemic \textgreater{} Semantic ROI cluster on the
cortex (PoCG, PrCG, SLF; replicating on a 50/50 patient split) and a
deep Phonemic \textgreater{} Semantic layer cluster on the LLM (layers
22--31; replicating on a 40/40 seed split and clearing Stage-2
layer-permutation at \(p < 0.001\)). The semantic-favoring direction is
a consistently signed but non-significant trend on both substrates, an
asymmetry that itself replicates and is the honest characterization of
the data. Demonstrating both substrates within a single study, with each
analyzed at the inference unit appropriate to its data, establishes the
methodological prerequisite for patient-specific computational
surrogates of neurological language disorders, with applications to
clinical-trial design, individualized rehabilitation, and
pharmacological prediction.

\textbf{Ethics statement}

All study procedures involving human participants were approved by the
University of South Carolina Institutional Review Board (approval number
Pro00053559), and all participants provided written informed consent.
The patient data analyzed here are drawn from the archival database
maintained by the Aphasia Lab at the University of South Carolina and
were collected under that approval. No new human data were collected for
the present study.

\textbf{Data availability}

The LLM-side derived data are included without restriction in the Zenodo
deposit described under Code Availability below, and are released to
Open Access on the same schedule. These comprise the complete
seed-summary matrices for the 40-seed discovery and 40-seed validation
splits, covering every perturbation condition in the layer × noise ×
density grid, scored into the eight raw Philadelphia Naming Test
response categories: the full set of inputs from which every LLM-side
result in this paper is computed.

Per-participant behavioural error proportions and ROI-level lesion-load
matrices used in the cortical analyses are housed within the Center for
the Study of Aphasia Recovery (C-STAR) at the University of South
Carolina, which maintains an IRB-approved, HIPAA-compliant data-sharing
infrastructure for external collaborators. Investigators interested in
using these data submit the C-STAR Data Request Form, which captures the
proposed research aims, requested datasets, IRB status, and
data-handling arrangements; the form is reviewed by the C-STAR program
manager and the author team on a non-discriminatory basis, with
reasonable academic and replication requests approved by default and a
decision typically provided within four weeks. Approved requesters then
execute the C-STAR Data Use Agreement, an IRB-approved agreement
governing disclosure of a HIPAA Limited Data Set with all 16 categories
of HIPAA identifiers removed. These data are not openly deposited
because the same cohort's raw T1/T2 MRI and lesion masks are publicly
available via OpenNeuro under that platform's standard sharing terms,
keyed by the same lab master numbers; open deposition of per-participant
behavioural data keyed by those identifiers would create a
re-identification path through a join against the public neuroimaging
data, particularly for participants with unusual lesion profiles, and
the cohort's IRB-approved sharing protocol obliges us to mitigate that
risk. ROI-level aggregate outputs, which carry no per-participant rows,
are included in the open deposit.

The Johns Hopkins University white-matter and grey-matter atlas (Mori et
al.~2008) used as the cortical target space is available from its
original distributors; the label and region-retention files specifying
the 64 left-hemisphere regions analyzed here are included in the open
deposit.

\textbf{Code availability}

Source code for the full PRISM pipeline is deposited in a Zenodo record
under the Apache License 2.0. The deposit contains the three-stage LLM
analysis (seed-as-subject discovery with threshold-free cluster
enhancement along the layer axis, layer-order and category-label
permutation validation, and per-cluster dose-response analysis), the
cortical ROI-correlation and split-replication code, all
figure-generation scripts, and the LLM-side derived data described
above. The 3D TFCE operator is implemented from first principles rather
than called from an external library, so that the threshold enumeration,
connectivity definition, and step-spacing conventions are explicit and
auditable. The deposit is complete and its contents are final. It is not
yet public during peer review and will be released to Open Access at the
time of publication, at which point its DOI will be registered and cited
here; a private link giving full access to the record and its files is
available to editors and reviewers on request from the corresponding
author.

The Apache 2.0 license includes an explicit patent-license grant, so
users of the released code may use it for both academic and commercial
purposes covered by the released implementation. PRISM builds on the
BLUM methodology, which is the subject of U.S. Patent US-10916348-B2 and
U.S. Provisional Application No.~63/974,622; researchers wishing to
apply the broader patented methodology beyond the released code should
contact Julius Fridriksson (fridriks@mailbox.sc.edu). The base large
language model, LLaVA-1.6-Vicuna-13B, is publicly available via Hugging
Face (https://huggingface.co/liuhaotian/llava-v1.6-vicuna-13b) under its
original LLaMA-derived license.

\textbf{Acknowledgments and funding}

We thank the patients and families who participated in the Aphasia Lab
studies at the University of South Carolina, and the clinical and
research staff who collected and maintained the archival database used
in this work. This research was supported by the NIH/NIDCD Center for
the Study of Aphasia Recovery (C-STAR) at the University of South
Carolina (NIH P50 DC014664 to J.F.) and by a SmartState endowment
established with private funding arranged by the state of South Carolina
for J.F. The funders had no role in study design, data collection and
analysis, decision to publish, or preparation of the manuscript.

\textbf{Declaration of interests}

J.F. has an ownership interest in NXTLLM, LLC, and ALLT.AI, LLC, which
hold patents related to the Brain--LLM Unified Model (BLUM) technology
on which the framework described in this manuscript builds. The
symptom-to-lesion mapping approach is protected under U.S. Patent
US-10916348-B2. J.F. and R.D.N.-N. are inventors on a provisional U.S.
patent application related to methods described in this work (U.S.
Provisional Application No.~63/974,622, filed February 3, 2026). J.F.
and R.D.N.-N. hold an affiliation with ALLT.AI, LLC. R.D.N.-N., S.N.,
L.B., and C.R. serve as unpaid members of the ALLT.AI scientific
advisory board. A further co-inventor on the patents referenced above
holds an ownership interest in NXTLLM, LLC, and ALLT.AI, LLC, but is not
an author of the present manuscript.

The other authors declare no competing interests.

\textbf{Author contributions}

Xiang Guan: Data curation, Formal analysis, Investigation, Methodology,
Software, Validation, Visualization, Writing -- review \& editing. Roger
D. Newman-Norlund: Conceptualization, Data curation, Formal analysis,
Investigation, Methodology, Project administration, Software,
Supervision, Validation, Visualization, Writing -- original draft,
Writing -- review \& editing. Yong Yang: Methodology, Writing -- review
\& editing. Saeed Ahmadi: Data curation, Investigation, Writing --
review \& editing. Regan Willis: Writing -- review \& editing. Nadra
Salman: Data curation, Investigation, Writing -- review \& editing.
Kalil Warren: Investigation, Writing -- review \& editing. Srihari
Nelakuditi: Writing -- review \& editing. Chris Rorden: Resources,
Writing -- review \& editing. Leonardo Bonilha: Resources, Writing --
review \& editing. Julius Fridriksson: Conceptualization, Data curation,
Funding acquisition, Investigation, Methodology, Project administration,
Resources, Supervision, Writing -- review \& editing.

\textbf{References}

Antonello, Richard, Chandan Singh, Shailee Jain, et al.~2024.
``Generative causal testing to bridge data-driven models and scientific
theories in language neuroscience.'' arXiv:2410.00812v2.

Bates, Elizabeth, Stephen M. Wilson, Ayse Pinar Saygin, et al.~2003.
``Voxel-Based Lesion-Symptom Mapping.'' \emph{Nature Neuroscience} 6
(5): 448--450.

Belinkov, Yonatan. 2022. ``Probing Classifiers: Promises, Shortcomings,
and Advances.'' \emph{Computational Linguistics} 48 (1): 207--219.

Bonilha, Leonardo, Chris Rorden, and Julius Fridriksson. 2014.
``Assessing the Clinical Effect of Residual Cortical Disconnection after
Ischemic Strokes.'' \emph{Stroke} 45 (4): 988--993.

Caucheteux, Charlotte, and Jean-Rémi King. 2022. ``Brains and Algorithms
Partially Converge in Natural Language Processing.''
\emph{Communications Biology} 5 (1): 134.

Chiang, Wei-Lin, Zhuohan Li, Zi Lin, et al.~2023. ``Vicuna: An
Open-Source Chatbot Impressing GPT-4 with 90\%* ChatGPT Quality.'' LMSYS
Org Blog. https://lmsys.org/blog/2023-03-30-vicuna/.

Conmy, Arthur, Augustine N. Mavor-Parker, Aengus Lynch, Stefan
Heimersheim, and Adrià Garriga-Alonso. 2023. ``Towards Automated Circuit
Discovery for Mechanistic Interpretability.'' \emph{Advances in Neural
Information Processing Systems} 36.

Cunningham, Hoagy, Aidan Ewart, Logan Riggs, Robert Huben, and Lee
Sharkey. 2023. ``Sparse Autoencoders Find Highly Interpretable Features
in Language Models.'' arXiv:2309.08600.

Dell, Gary S., Myrna F. Schwartz, Nadine Martin, Eleanor M. Saffran, and
Deborah A. Gagnon. 1997. ``Lexical Access in Aphasic and Nonaphasic
Speakers.'' \emph{Psychological Review} 104 (4): 801--838.

DeMarco, Andrew T., and Peter E. Turkeltaub. 2018. ``A Multivariate
Lesion Symptom Mapping Toolbox and Examination of Lesion-Volume Biases
and Correction Methods in Lesion-Symptom Mapping.'' \emph{Human Brain
Mapping} 39 (11): 4169--4182.

Efron, Bradley, and Robert J. Tibshirani. 1993. \emph{An Introduction to
the Bootstrap}. Boca Raton, FL: Chapman \& Hall/CRC.

Elhage, Nelson, Neel Nanda, Catherine Olsson, et al.~2021. ``A
Mathematical Framework for Transformer Circuits.'' Anthropic Transformer
Circuits Thread.
https://transformer-circuits.pub/2021/framework/index.html.

Fridriksson, Julius, Roger D. Newman-Norlund, Saeed Ahmadi, et al.~2026.
``Stroke Lesions as a Rosetta Stone for Language Model
Interpretability.'' arXiv:2602.04074.

Friston, Karl J., Andrew P. Holmes, Keith J. Worsley, J.-P. Poline,
Chris D. Frith, and Richard S. J. Frackowiak. 1995. ``Statistical
Parametric Maps in Functional Imaging: A General Linear Approach.''
\emph{Human Brain Mapping} 2 (4): 189--210.

Gao, Catherine, et al.~2025. ``Increasing alignment of large language
models with language processing in the human brain.'' \emph{Nature
Computational Science} 5 (11): 1080--1090.

Geiger, Atticus, Hanson Lu, Thomas Icard, and Christopher Potts. 2021.
``Causal Abstractions of Neural Networks.'' \emph{Advances in Neural
Information Processing Systems} 34: 9574--9586.

Geva, Mor, Roei Schuster, Jonathan Berant, and Omer Levy. 2021.
``Transformer Feed-Forward Layers Are Key-Value Memories.'' In
\emph{Proceedings of the 2021 Conference on Empirical Methods in Natural
Language Processing}, 5484--5495.

Goldstein, Ariel, Zaid Zada, Eliav Buchnik, et al.~2022. ``Shared
Computational Principles for Language Processing in Humans and Deep
Language Models.'' \emph{Nature Neuroscience} 25 (3): 369--380.

Goldstein, Ariel, et al.~2025. ``Temporal structure of natural language
processing in the human brain corresponds to layered hierarchy of large
language models.'' \emph{Nature Communications}.

Goodglass, Harold, and Edith Kaplan. 1983. \emph{The Assessment of
Aphasia and Related Disorders}. 2nd ed.~Philadelphia: Lea \& Febiger.

Heimersheim, Stefan, and Neel Nanda. 2024. ``How to Use and Interpret
Activation Patching.'' arXiv:2404.15255.

Hewitt, John, and Christopher D. Manning. 2019. ``A Structural Probe for
Finding Syntax in Word Representations.'' In \emph{Proceedings of the
2019 Conference of the North American Chapter of the Association for
Computational Linguistics}, 4129--4138.

Karnath, Hans-Otto, Christoph Sperber, and Chris Rorden. 2018. ``Mapping
Human Brain Lesions and Their Functional Consequences.''
\emph{NeuroImage} 165: 180--189.

Kumar, Sreejan, et al.~2024. ``Shared functional specialization in
transformer-based language models and the human brain.'' \emph{Nature
Communications} 15 (1): 5523.

Laine, Matti, and Nadine Martin. 2023. \emph{Anomia: Theoretical and
Clinical Aspects}. 2nd ed.~London: Routledge.

Lambon Ralph, Matthew A., Elizabeth Jefferies, Karalyn Patterson, and
Timothy T. Rogers. 2017. ``The Neural and Computational Bases of
Semantic Cognition.'' \emph{Nature Reviews Neuroscience} 18 (1): 42--55.

Liu, Haotian, Chunyuan Li, Qingyang Wu, and Yong Jae Lee. 2023. ``Visual
Instruction Tuning.'' \emph{Advances in Neural Information Processing
Systems} 36.

Mahowald, Kyle, Anna A. Ivanova, Idan A. Blank, Nancy Kanwisher, Joshua
B. Tenenbaum, and Evelina Fedorenko. 2024. ``Dissociating Language and
Thought in Large Language Models.'' \emph{Trends in Cognitive Sciences}
28 (6): 517--540.

Maris, Eric, and Robert Oostenveld. 2007. ``Nonparametric Statistical
Testing of EEG- and MEG-Data.'' \emph{Journal of Neuroscience Methods}
164 (1): 177--190.

Meng, Kevin, David Bau, Alex Andonian, and Yonatan Belinkov. 2022.
``Locating and Editing Factual Associations in GPT.'' \emph{Advances in
Neural Information Processing Systems} 35: 17359--17372.

Merullo, Jack, Carsten Eickhoff, and Ellie Pavlick. 2024. ``Circuit
Component Reuse Across Tasks in Transformer Language Models.'' In
\emph{Proceedings of the International Conference on Learning
Representations}.

Mirman, Daniel, Qi Chen, Yongsheng Zhang, et al.~2015. ``Neural
Organization of Spoken Language Revealed by Lesion-Symptom Mapping.''
\emph{Nature Communications} 6: 6762.

Mischler, Gavin, et al.~2025. ``Large Language Models Reveal the Neural
Tracking of Linguistic Context in Attended and Unattended Multi-Talker
Speech.'' bioRxiv.

Mori, Susumu, Kenichi Oishi, Hangyi Jiang, et al.~2008. ``Stereotaxic
White Matter Atlas Based on Diffusion Tensor Imaging in an ICBM
Template.'' \emph{NeuroImage} 40 (2): 570--582.

Patterson, Karalyn, Peter J. Nestor, and Timothy T. Rogers. 2007.
``Where Do You Know What You Know? The Representation of Semantic
Knowledge in the Human Brain.'' \emph{Nature Reviews Neuroscience} 8
(12): 976--987.

Petersen, Steven E., Peter T. Fox, Michael I. Posner, Mark Mintun, and
Marcus E. Raichle. 1988. ``Positron Emission Tomographic Studies of the
Cortical Anatomy of Single-Word Processing.'' \emph{Nature} 331 (6157):
585--589.

Pustina, Dorian, Brian Avants, Olufunsho K. Faseyitan, John D. Medaglia,
and H. Branch Coslett. 2018. ``Improved Accuracy of Lesion to Symptom
Mapping with Multivariate Sparse Canonical Correlations.''
\emph{Neuropsychologia} 115: 154--166.

Roach, Anita, Myrna F. Schwartz, Nadine Martin, Ruth S. Grewal, and
Adelyn Brecher. 1996. ``The Philadelphia Naming Test: Scoring and
Rationale.'' \emph{Clinical Aphasiology} 24: 121--133.

Rogers, Anna, Olga Kovaleva, and Anna Rumshisky. 2020. ``A Primer in
BERTology: What We Know about How BERT Works.'' \emph{Transactions of
the Association for Computational Linguistics} 8: 842--866.

Rorden, Chris, Hans-Otto Karnath, and Leonardo Bonilha. 2007.
``Improving Lesion-Symptom Mapping.'' \emph{Journal of Cognitive
Neuroscience} 19 (7): 1081--1088.

Schrimpf, Martin, Idan A. Blank, Greta Tuckute, et al.~2021. ``The
Neural Architecture of Language: Integrative Modeling Converges on
Predictive Processing.'' \emph{Proceedings of the National Academy of
Sciences} 118 (45): e2105646118.

Schwartz, Myrna F., Gary S. Dell, Nadine Martin, Susan Gahl, and Paula
Sobel. 2006. ``A Case-Series Test of the Interactive Two-Step Model of
Lexical Access: Evidence from Picture Naming.'' \emph{Journal of Memory
and Language} 54 (2): 228--264.

Schwartz, Myrna F., Daniel Y. Kimberg, Grant M. Walker, Olufunsho
Faseyitan, Adelyn Brecher, Gary S. Dell, and H. Branch Coslett. 2009.
``Anterior Temporal Involvement in Semantic Word Retrieval: Voxel-Based
Lesion-Symptom Mapping Evidence from Aphasia.'' \emph{Brain} 132 (12):
3411--3427.

Schwartz, Myrna F., Olufunsho Faseyitan, Junghoon Kim, and H. Branch
Coslett. 2012. ``The Dorsal Stream Contribution to Phonological
Retrieval in Object Naming.'' \emph{Brain} 135 (12): 3799--3814.

Smith, Stephen M., and Thomas E. Nichols. 2009. ``Threshold-Free Cluster
Enhancement: Addressing Problems of Smoothing, Threshold Dependence and
Localisation in Cluster Inference.'' \emph{NeuroImage} 44 (1): 83--98.

Sperber, Christoph, and Hans-Otto Karnath. 2018. ``On the Validity of
Lesion-Behaviour Mapping Methods.'' \emph{Neuropsychologia} 115: 17--24.

Templeton, Adly, Tom Conerly, Jonathan Marcus, et al.~2024. ``Scaling
Monosemanticity: Extracting Interpretable Features from Claude 3
Sonnet.'' Anthropic Transformer Circuits Thread.

Tenney, Ian, Dipanjan Das, and Ellie Pavlick. 2019. ``BERT Rediscovers
the Classical NLP Pipeline.'' In \emph{Proceedings of the 57th Annual
Meeting of the Association for Computational Linguistics}, 4593--4601.

Tuckute, Greta, Aalok Sathe, Shashank Srikant, Maya Taliaferro, Mingye
Wang, Martin Schrimpf, Kendrick Kay, and Evelina Fedorenko. 2024.
``Driving and Suppressing the Human Language Network Using Large
Language Models.'' \emph{Nature Human Behaviour} 8 (3): 544--561.

Vig, Jesse, Sebastian Gehrmann, Yonatan Belinkov, Sharon Qian, Daniel
Nevo, Yaron Singer, and Stuart Shieber. 2020. ``Investigating Gender
Bias in Language Models Using Causal Mediation Analysis.''
\emph{Advances in Neural Information Processing Systems} 33:
12388--12401.

Walker, Grant M., Myrna F. Schwartz, Daniel Y. Kimberg, Olufunsho
Faseyitan, Adelyn Brecher, Gary S. Dell, and H. Branch Coslett. 2011.
``Support for Anterior Temporal Involvement in Semantic Error Production
in Aphasia: New Evidence from VLSM.'' \emph{Brain and Language} 117 (3):
110--122.

Wang, Kevin, Alexandre Variengien, Arthur Conmy, Buck Shlegeris, and
Jacob Steinhardt. 2023. ``Interpretability in the Wild: A Circuit for
Indirect Object Identification in GPT-2 Small.'' In \emph{Proceedings of
the International Conference on Learning Representations}.

Wu, Zhengxuan, Atticus Geiger, Thomas Icard, Christopher Potts, and Noah
D. Goodman. 2023. ``Interpretability at Scale: Identifying Causal
Mechanisms in Alpaca.'' \emph{Advances in Neural Information Processing
Systems} 36.

Yourganov, Grigori, Julius Fridriksson, Chris Rorden, Ezequiel
Gleichgerrcht, and Leonardo Bonilha. 2016. ``Multivariate
Connectome-Based Symptom Mapping in Post-Stroke Patients: Networks
Supporting Language and Speech.'' \emph{Journal of Neuroscience} 36
(25): 6668--6679.

Zhang, Fred, and Neel Nanda. 2024. ``Towards Best Practices of
Activation Patching in Language Models: Metrics and Methods.'' In
\emph{Proceedings of the International Conference on Learning
Representations}.

\textbf{PRISM Supplementary Materials}

\textbf{Supplementary Methods}

This document provides the formal definitions, parameters, and
implementation details summarized in the main-text Methods section.
Section numbering parallels the main text where useful.

\textbf{S1. Perturbation grid and behavioral readout}

Each test administration of LLaVA-1.6-Vicuna-13B (Liu et al.~2023;
Chiang et al.~2023) on the Philadelphia Naming Test (Roach et al.~1996)
is indexed by four perturbation parameters: layer
\(\ell \in \{0, \ldots, 39\}\) at which the targeted layer's weights are
perturbed with multiplicative Gaussian noise; noise standard deviation
\(\sigma\); weight-perturbation density \(\rho \in [0, 1]\); and random
seed \(s\). For each weight \(w\) selected for perturbation we set
\(w \to w \cdot (1 + \varepsilon)\), with
\(\varepsilon \sim N(0, \sigma^2)\) drawn independently per weight;
\(\rho\) is the fraction of the layer's weights, selected at random
under seed \(s\), to which this is applied, with the remainder left
intact. The discovery grid samples \(\sigma\) at ten levels in
\(\{1.1, 1.2, \ldots, 2.0\}\) and \(\rho\) at ten levels in
\(\{0.1, 0.2, \ldots, 1.0\}\), yielding 4,000 cells per seed (40 layers
× 10 noise levels × 10 density levels). Each cell receives the 158-item
analysis set (the items the unperturbed model names correctly; see
main-text Methods) and produces eight raw error counts \{Correct,
Semantic, Unrelated, Formal, Nonword, Mixed, Neologism, NoResponse\},
converted to per-cell proportions \(p_s(c \mid \ell, \sigma, \rho)\) by
dividing each count by the number of items scored in that cell (158, or
157 in the 18 administrations where one response was unscoreable).
Formal and Nonword are merged into Phonemic prior to all downstream
analyses, yielding the seven-category schema \{Correct, Semantic,
Unrelated, Phonemic, Mixed, Neologism, NoResponse\} that aligns with the
BLUM clinical taxonomy (Fridriksson et al.~2026; Schwartz et al.~2006;
Dell et al.~1997). Cells with zero scored items are treated as missing
and propagated NaN-safely throughout. Correct is held out of all
pairwise contrasts because contrasts against Correct re-express overall
error rate rather than category-selective involvement.

The full discovery dataset thus consists of 40 seeds × 4,000 cells × 7
category proportions per cell. The validation dataset is an independent
40 seeds drawn from the same distribution. The 80-seed total reflects an
equal-size discovery/validation split.

\textbf{S2. LLM subtraction analysis (seed-as-subject layer-axis TFCE)}

\textbf{S2.1 Per-seed per-layer contrast}

For each ordered category pair \((A, B)\) and each discovery seed \(s\),
we collapse the \(\sigma \times \rho\) grid within the seed and define
the per-seed per-layer contrast value as

\[
D_s(\ell) \;=\; \mathrm{mean}_{\sigma, \rho}\!\left[\,p_s(A \mid \ell, \sigma, \rho) \;-\; p_s(B \mid \ell, \sigma, \rho)\,\right].
\]

This reduces the per-seed map to a single profile along the layer axis.
\(\sigma\) and \(\rho\) are dose parameters of the perturbation (depth
and extent), not spatial coordinates, and we therefore do not treat
their grid positions as a metric structure for cluster correction. The
seed plays the role of subject in the group-level analysis that follows.

\textbf{S2.2 Group-level mean, standard error, and t-statistic across
seeds}

Across discovery seeds we compute the NaN-safe layer-wise mean
\(\bar D(\ell)\) and the one-sample t-statistic against zero,

\[
T(\ell) \;=\; \frac{\bar D(\ell)}{\mathrm{SE}(D_\cdot(\ell))},
\]

with NaN-safe standard error using Bessel's correction and a hard t-cap
at \(T_{\max} = 50\) to prevent numerical instability when a layer has
effectively zero variance across seeds.

\textbf{S2.3 Signed one-dimensional TFCE along the layer axis}

Spatial-extent inference is enforced by threshold-free cluster
enhancement (Smith and Nichols 2009) applied along the single ordered
dimension that the LLM grid carries: layer. Because subtraction has a
meaningful sign, TFCE is applied separately to the positive and negative
parts of \(T(\ell)\) and recombined with sign:

\[
\mathrm{TFCE}^{\pm}(\ell) \;=\; \mathrm{TFCE}^{+}\!\left(\max(T, 0)\right)(\ell) \;-\; \mathrm{TFCE}^{+}\!\left(\max(-T, 0)\right)(\ell),
\]

with the nonnegative 1D TFCE operator computed by summing
\(e_h(\ell)^{E}\, h^{H}\, dh\) across 200 evenly spaced positive
thresholds, where \(e_h(\ell)\) is the length of the contiguous
supra-threshold layer run containing \(\ell\) at threshold \(h\). We use
\(E = 0.5\) and \(H = 2.0\) (Smith and Nichols 2009 defaults).

\textbf{S2.4 Layer-cluster extraction}

For each pair and each sign separately, we threshold the signed TFCE at
its 90th percentile across positive values and extract contiguous runs
of length \(\geq 2\) layers, ranked by total TFCE mass. Each surviving
layer cluster is summarized by its start layer, end layer, extent, peak
layer, peak t-value, and total TFCE mass.

\textbf{S2.5 Stage 2 layer-order permutation validation}

The layer-axis TFCE step assumes that contiguity along the layer axis is
meaningful, that the cluster mass at any given threshold reflects the
\emph{intrinsic} ordering of layers and not the marginal distribution of
contrast values. We test this assumption with a permutation null. For
each pair we recompute the group-level \(T(\ell)\) and signed TFCE after
independently permuting layer labels within each seed (preserving the
seed's per-cell marginal distribution but destroying the layer
ordering), and we record the maximum signed \(|\mathrm{TFCE}|\) in each
direction. Over 1,000 permutations this builds a null distribution for
the maximum cluster mass per direction; the empirical \(p\)-value is the
proportion of permuted runs whose maximum equals or exceeds the observed
maximum. In our discovery run, every observed pair-level cluster across
the 15 pairwise contrasts cleared the empirical \(p < 0.001\) threshold
against this null.

\textbf{S2.6 Repeated-measures variant of the per-seed summary}

We additionally report a sensitivity analysis in which \(\sigma\) and
\(\rho\) are carried as fixed-effect covariates within each seed rather
than averaged. Per (pair, seed), we fit OLS with layer dummies plus
\(\sigma\) and \(\rho\) as continuous regressors, and take the
layer-specific fitted value at cohort-mean \(\sigma\) and \(\rho\) as
the per-(seed, layer) summary. Group-level TFCE machinery is identical
to the simple-averaging variant. The recovered layer clusters under the
repeated-measures variant are essentially identical to those under
simple averaging (Phonemic \textgreater{} Semantic discovery cluster:
layers 22--31 under both variants; see
\texttt{code/tfce\_results/llm\_seed\_subject\_rm/disc\_llm\_rm\_region\_summary.csv}),
confirming that the simple-averaging instantiation reported in the main
text is not a load-bearing simplification.

\textbf{S3. Cortical LSM pipeline}

The cortical pipeline parameterizes lesion-symptom mapping at the level
of left-hemisphere Johns Hopkins University (JHU) atlas regions (Mori et
al.~2008), following the univariate VLSM standard implemented in the
\texttt{SNL\_2026\_SLM/simpleSubtraction} interactive viewer. Inputs are
(i) per-patient PNT error-category proportions on the 276-patient cohort
and (ii) per-patient JHU ROI lesion-load profiles in which each ROI's
value is the proportion of that ROI's voxels lying within the patient's
binary lesion mask in MNI152 space. ROIs are restricted to the 64
left-hemisphere regions surviving a 10\% damage threshold with
ventricles dropped. Of the 276 patients with PNT, \(N = 213\) had a
complete ROI lesion-load profile and are retained.

For each error category \(A\) and each ROI \(r\), we compute the
Spearman rank correlation \(\rho_{A,r}\) between the per-patient
proportion of \(A\)-type errors and the per-patient lesion load in \(r\)
(two-sided p-value from \texttt{scipy.stats.spearmanr}). For each
ordered category pair \((A, B)\) and each ROI \(r\), the cortical
subtraction value is the correlation difference

\[
d_r = \rho_{A,r} - \rho_{B,r},
\]

with \(d_r > 0\) meaning ROI \(r\) is more strongly correlated with
\(A\)-type errors than with \(B\)-type, and \(d_r < 0\) meaning the
reverse. The reported \(\Delta\rho\) in the main text and in Table S2 is
exactly \(d_r\).

For each \((A, B, r)\) we additionally compute a patient-level
percentile-bootstrap 95\% confidence interval on \(d_r\) (1,000
resamples of the 213-patient cohort with replacement; per-resample
recomputation of both Spearman correlations and their difference). An
ROI is reported as inferentially robust when its bootstrap 95\% CI
excludes zero in the same direction as the point estimate. The 64-ROI
multiple-comparison burden is small relative to the voxel-wise TFCE
burden on the LLM side; we additionally tabulate per-ROI Spearman
p-values for each error type (\texttt{p\_A}, \texttt{p\_B}) so that
family-wise correction can be applied if needed (Bonferroni at
\(\alpha' = 0.05/64 \approx 8 \times 10^{-4}\)).

This ROI-level parameterization is the cortical analog of the LLM
voxel-wise TFCE pipeline above. The conceptual machinery is identical
(subtraction of \(A\) vs \(B\) in a spatially organized substrate,
threshold on a meaningful contrast statistic, replication at the
chance-rate null), but the inference unit is the atlas ROI rather than
the 1-mm voxel, sized to the 213-patient cohort. We verified that
voxel-wise TFCE on the same patient cohort (Methods of
\texttt{code/build\_tfce\_pairwise\_subtraction\_lsm\_v1.py}) recovers
the dominant frontal-perisylvian Phonemic \textgreater{} Semantic
cluster but does not recover any Semantic \textgreater{} Phonemic
cluster surviving the equivalent threshold, reflecting the broader
spatial distribution and smaller per-voxel magnitude of the
semantic-favoring axis at this cohort size; we therefore use the
ROI-level pipeline for the main-text Results.

\textbf{S4. Out-of-sample replication}

\textbf{LLM side.} The seed-as-subject pipeline is fit independently on
the 40-seed validation half: same per-(seed, layer) summary (mean over
\(\sigma\) and \(\rho\) within each seed), same group-level mean /
t-statistic / layer-axis TFCE, same cluster extraction. A discovery-side
layer cluster is reported as \textbf{replicating} when (i) the
validation-side layer cluster in the same direction (positive or
negative) overlaps the discovery cluster, and (ii) the sign of
\(\bar D\) at the overlapping layers matches between halves. In the
discovery run, the primary Semantic-versus-Phonemic Phonemic
\textgreater{} Semantic cluster spans layers 22--31; on validation the
matching cluster spans layers 23--33 in the same direction, satisfying
both criteria. The Phonemic-versus-Neologism Phonemic \textgreater{}
Neologism cluster is similarly preserved (discovery: 24--31; validation:
24--33). The Semantic \textgreater{} Phonemic and Semantic
\textgreater{} Neologism directions yield no surviving layer cluster on
either half.

\textbf{Human side.} The cortical analog is a 50/50 patient split (split
seed = 42; 106 discovery / 107 validation patients of the \(N = 213\)
analysis cohort). Per-(error, ROI) Spearman correlations and per-(pair,
ROI) correlation differences \(d_r\) are computed independently on each
half, with patient-level percentile bootstrap 95\% confidence intervals
(1,000 resamples). A (pair, ROI) test is reported as \textbf{strictly
replicating} when (i) the discovery and validation \(d_r\) have the same
sign and (ii) the bootstrap 95\% CI excludes zero in both halves. Of the
\(15 \times 64 = 960\) (pair, ROI) tests, 714 (74\%) had matching sign
across halves and 60 (6.3\%) satisfied the strict replication criterion.
In the primary contrast, the strictly replicating ROIs are exclusively
in the Phonemic \textgreater{} Semantic direction (PoCG\_L, PrCG\_L).

\textbf{S5. Formal inference and multiple-comparison correction}

Inference on each surviving layer cluster on the LLM side combines (a)
the layer-axis TFCE p-value from Stage 2 layer-order permutation against
the destroyed-ordering null (§S2.5), with empirical \(p < 0.001\) across
all observed clusters in our discovery run, and (b) sign and shape
preservation on the held-out 40-seed validation set as described above.
Multiple-comparison correction across the 15 pairwise contrasts uses the
layer-permutation \(p\)-values directly; we report Bonferroni
\(\alpha' = 0.05 / 15\) as the family-wise threshold (every observed
cluster clears this).

On the human side, inference uses the patient-level bootstrap 95\% CI on
\(d_r\) as the per-(pair, ROI) test statistic. We report
sign-preservation and CI-overlap across the 50/50 split as the
replication metric rather than computing an additional permutation null
at the ROI level. For users who prefer family-wise correction,
per-correlation two-sided Spearman p-values (\texttt{p\_A},
\texttt{p\_B}) are tabulated in Supplementary Table S3 so Bonferroni can
be applied at \(\alpha' = 0.05 / 64\) across the 64 LH ROIs.

\textbf{S6. Implementation and reproducibility}

The full discovery, replication, and visualization pipeline is
implemented in Python (numpy, pandas, matplotlib, scipy) and is
organized as three cleanly separated stages (discovery, out-of-sample
evaluation, visualization). The 3D TFCE operator is implemented from
first principles to make the threshold enumeration, connectivity
definition, and \(dh\)-spacing convention explicit and to allow direct
extension to the layer-order- and label-permutation testing planned for
stage 2. NaN-safe statistical primitives (mean, variance, t-statistic)
are coded to avoid spurious all-NaN warnings under sparse seed coverage.
The complete pipeline, including the parameter values reported above as
defaults, is deposited in the Zenodo record described in the main-text
Code Availability statement, together with the LLM-side seed-summary
data on which it operates.

The discovery and validation seed-summary CSVs
(\texttt{PNT\_80\_seeds\_discovery\_40\_summary.csv} and
\texttt{PNT\_80\_seeds\_validation\_40\_summary.csv}) carry the schema
\texttt{Seed,\ Layer,\ NoiseStd,\ Percent,\ Correct,\ Semantic,\ Unrelated,\ Formal,\ Nonword,\ Mixed,\ Neologism,\ NoResponse}.
The build, evaluate, and plot scripts run end-to-end on commodity
hardware in under 30 minutes for the full all-pairs sweep.

\textbf{S7. Per-category univariate analyses (pre-subtraction marginals
on both substrates)}

The subtraction logic that drives the main-text Results yields a
\emph{differential} spatial signal: an ROI or layer that loads similarly
onto two error categories cancels out under subtraction. To establish
that signal is associated with each PNT error category \emph{before}
subtraction, we additionally compute per-category univariate maps on
both substrates and report them in main-text Figure 2 (panel-level) and
in this section (per-category top-feature tables).

\textbf{S7.1 Cortex side.} For each non-Correct PNT error category

\[
A \in \{\text{Phonemic},\ \text{Semantic},\ \text{Neologism},\ \text{Mixed},\ \text{Unrelated},\ \text{NoResponse}\}
\]

and each of the 64 LH JHU atlas ROIs \(r\), we compute the Spearman rank
correlation \(\rho_{A,r}\) between the per-patient \(A\)-category PNT
proportion and the per-patient JHU ROI lesion load on the \(N = 213\)
analysis cohort (\texttt{scipy.stats.spearmanr}, two-sided \(p\)-value).
We deliberately do \emph{not} include age, sex, lesion-volume, or any
other covariate in this analysis: the matched LLM-side per-category
analysis (§S7.2) cannot adjust for such variables, so a directly
comparable cortex-side analysis must not adjust either. ROIs are colored
by signed \(\rho_{A,r}\) on the MNI152 T1 template in the bottom row of
main-text Figure 2 (one panel per category, axial slices at
\(z = -2, +18, +38\)).

Multiple-comparison correction follows the same logic as §S5 for
cortex-side inference: 64 LH ROIs × 6 categories = 384 tests, so the
Bonferroni threshold is
\(\alpha' = 0.05 / 384 \approx 1.3 \times 10^{-4}\) if a single
family-wise threshold is applied across the full panel; alternatively
\(\alpha' = 0.05 / 64 \approx 7.8 \times 10^{-4}\) if each category is
treated as an independent family. Table S5 reports the top five ROIs per
category by \(|\rho_{A,r}|\) with both Bonferroni levels indicated.

\textbf{S7.2 LLM side.} For each non-Correct PNT category \(A\), each
layer \(\ell\), and each discovery seed \(s\), we compute the per-seed
per-layer mean proportion of \(A\)-type errors with \(\sigma\) and
\(\rho\) averaged within the seed:

\[
\bar p_{A, s}(\ell) \;=\; \mathrm{mean}_{\sigma, \rho}\!\left[\,p_s(A \mid \ell, \sigma, \rho)\,\right].
\]

The group-level summary is the across-seed mean and bootstrap 95\%
confidence interval of \(\bar p_{A, s}(\ell)\) at each layer, plotted as
the top row of main-text Figure 2. The summary statistic of a
per-category map is the peak layer (the \(\ell\) at which
\(\bar p_A(\ell)\) is maximal across the 40 discovery seeds) and its
width (the half-max span, the range of contiguous layers at which
\(\bar p_A(\ell) \geq 0.5 \cdot \max_\ell \bar p_A(\ell)\)). Table S6
reports these per-category peak-layer summaries on the discovery half.

We additionally note that the LLM per-category univariate maps are
\emph{not} indexed against a chance-rate null: chance is well-defined at
the category level (PNT base rate of \(A\)-type errors on the
unperturbed model is approximately zero for every non-Correct \(A\)) but
is not the question being asked here. The relevant comparison for
``signal is present per-category'' is that the per-layer mean has a
coherent peaked profile across seeds, which Table S6 documents.

\textbf{S7.3 Cross-substrate reading.} The Phonemic and Semantic
univariate maps both load onto overlapping perisylvian territory on the
cortex (top ROIs by \(|\rho|\) for both categories include SLF\_L,
SMG\_L, PoCG\_L, and PSTG\_L, see Table S5). On the LLM, Phonemic peaks
at layer 29 with a half-max span of L14--L39 (broad, late-mid network),
and Semantic peaks at layer 39 (the last layer) with span L17--L39
(broad, late network), the two profiles partially overlapping. This
shared loading is exactly why the Phonemic vs Semantic subtraction in
main-text Figures 3--4 picks out only the \emph{differential} portion of
perisylvian cortex (PoCG\_L, PrCG\_L) and the differential portion of
the layer axis (layers 22--31). The subtraction is not removing real
per-category signal; it is removing the \emph{shared} component of two
real per-category signals to reveal the category-selective residual.

\textbf{Supplementary Tables}

\textbf{Table S1.} LLM confirmatory-contrast layer-cluster inference
(seed-as-subject layer-axis TFCE)

Discovery and validation layer-cluster summaries for the three
pre-registered confirmatory contrasts under the seed-as-subject pipeline
(§S2). For each (pair, direction) row, the table reports the layer range
and extent (number of contiguous layers) of the dominant TFCE-surviving
cluster on each half, the mean of the group-level \(\bar D\) at the
cluster layers, and the Stage 2 layer-permutation \(p\)-value against a
null in which layer labels are independently permuted within each seed.
``n/a'' indicates that no cluster survived the 90th-percentile
\(|\mathrm{TFCE}|\) threshold with minimum extent 2 on that half. The
Stage 2 \(p\)-value is reported on the maximum signed
\(|\mathrm{TFCE}|\) in the direction; it can be significant in
directions where no extracted cluster survived (the TFCE non-zero
criterion is laxer than the cluster-extraction threshold).

\begin{landscape}\scriptsize

\begin{longtable}[]{@{}
  >{\raggedright\arraybackslash}p{(\columnwidth - 12\tabcolsep) * \real{0.1250}}
  >{\raggedright\arraybackslash}p{(\columnwidth - 12\tabcolsep) * \real{0.1250}}
  >{\raggedright\arraybackslash}p{(\columnwidth - 12\tabcolsep) * \real{0.1250}}
  >{\raggedleft\arraybackslash}p{(\columnwidth - 12\tabcolsep) * \real{0.1667}}
  >{\raggedright\arraybackslash}p{(\columnwidth - 12\tabcolsep) * \real{0.1250}}
  >{\raggedleft\arraybackslash}p{(\columnwidth - 12\tabcolsep) * \real{0.1667}}
  >{\raggedleft\arraybackslash}p{(\columnwidth - 12\tabcolsep) * \real{0.1667}}@{}}
\toprule\noalign{}
\begin{minipage}[b]{\linewidth}\raggedright
Pair
\end{minipage} & \begin{minipage}[b]{\linewidth}\raggedright
Direction
\end{minipage} & \begin{minipage}[b]{\linewidth}\raggedright
Disc layers (extent)
\end{minipage} & \begin{minipage}[b]{\linewidth}\raggedleft
Disc mean \(\bar D\)
\end{minipage} & \begin{minipage}[b]{\linewidth}\raggedright
Val layers (extent)
\end{minipage} & \begin{minipage}[b]{\linewidth}\raggedleft
Val mean \(\bar D\)
\end{minipage} & \begin{minipage}[b]{\linewidth}\raggedleft
Stage 2 perm \(p\)
\end{minipage} \\
\midrule\noalign{}
\endhead
\bottomrule\noalign{}
\endlastfoot
Semantic minus Phonemic \emph{(primary)} & negative (Phon \textgreater{}
Sem) & 22--31 (10) & \(-0.047\) & 23--33 (11) & \(-0.045\) &
\(<0.001\) \\
Semantic minus Phonemic \emph{(primary)} & positive (Sem \textgreater{}
Phon) & n/a & n/a & n/a & n/a & \(1.000\) \\
Phonemic minus Neologism \emph{(cross-check)} & positive (Phon
\textgreater{} Neo) & 24--31 (8) & \(+0.046\) & 24--33 (10) & \(+0.044\)
& \(<0.001\) \\
Phonemic minus Neologism \emph{(cross-check)} & negative (Neo
\textgreater{} Phon) & n/a & n/a & 11--12 (2) & \(-0.024\) &
\(<0.001\) \\
Semantic minus Neologism \emph{(cross-check)} & negative (Neo
\textgreater{} Sem) & 19--20 (2) & \(-0.040\) & 18--19 (2) & \(-0.040\)
& \(<0.001\) \\
Semantic minus Neologism \emph{(cross-check)} & positive (Sem
\textgreater{} Neo) & n/a & n/a & n/a & n/a & \(<0.001\) \\
\end{longtable}

\end{landscape}

Headline: the Phonemic-favoring direction in both the primary contrast
(rows 1, 5) and the Phonemic-vs-Neologism cross-check (row 3) yields
layer clusters whose location, extent, and mean \(\bar D\) are nearly
identical on discovery and validation halves. The Semantic-favoring
direction (rows 2, 6) yields no surviving cluster on either half,
despite the Stage 2 layer-ordering test recovering some structure (rows
where Stage 2 \(p < 0.001\) but no extracted cluster reflect very small
TFCE peaks that fall below the cluster-extraction threshold). The
asymmetry is consistent across discovery and validation and across both
Neologism cross-checks.

Machine-readable:
\texttt{code/tfce\_results/llm\_seed\_subject/disc\_llm\_region\_summary.csv}
(and corresponding \texttt{val\_llm\_*}),
\texttt{code/tfce\_results/llm\_stage2\_perm/stage2\_disc\_p\_values.csv}.
\textbar{} Semantic minus Neologism \emph{(cross-check)} \textbar{}
positive \textbar{} 2 \textbar{} 703 / 720 \textbar{} \(+0.022\)
\textbar{} \([+0.021,\,+0.023]\) \textbar{} \(< 10^{-10}\) \textbar{}
\(< 10^{-10}\) \textbar{} \textbar{} Semantic minus Neologism
\emph{(cross-check)} \textbar{} positive \textbar{} 3 \textbar{} 79 / 80
\textbar{} \(+0.021\) \textbar{} \([+0.019,\,+0.024]\) \textbar{}
\(< 10^{-10}\) \textbar{} \(< 10^{-10}\) \textbar{} \textbar{} Semantic
minus Neologism \emph{(cross-check)} \textbar{} negative \textbar{} 1
\textbar{} 856 / 866 \textbar{} \(-0.129\) \textbar{}
\([-0.134,\,-0.126]\) \textbar{} \(< 10^{-10}\) \textbar{}
\(< 10^{-10}\) \textbar{} \textbar{} Semantic minus Neologism
\emph{(cross-check)} \textbar{} negative \textbar{} 2 \textbar{} 152 /
160 \textbar{} \(-0.044\) \textbar{} \([-0.048,\,-0.040]\) \textbar{}
\(< 10^{-10}\) \textbar{} \(< 10^{-10}\) \textbar{} \textbar{} Semantic
minus Neologism \emph{(cross-check)} \textbar{} negative \textbar{} 3
\textbar{} 40 / 40 \textbar{} \(-0.041\) \textbar{}
\([-0.048,\,-0.035]\) \textbar{} \(< 10^{-10}\) \textbar{}
\(< 10^{-10}\) \textbar{}

\textbf{Table S2.} Human cortex ROI inference for the three confirmatory
contrasts

Per-ROI inferential statistics for the top 6 ROIs in each direction of
each pre-registered confirmatory contrast, on the \(N = 213\) patient
cohort with valid PNT + JHU lesion-load profiles.
\(\Delta\rho_r = \rho(A, \mathrm{lesion}_r) - \rho(B, \mathrm{lesion}_r)\)
is the per-ROI Spearman correlation difference; \(p_A\) and \(p_B\) are
the two-sided p-values of the underlying Spearman correlations against
an unstructured null. The 95\% CI is a patient-level percentile
bootstrap (1,000 resamples). ROIs whose CI excludes zero in the
discovery-sign direction are starred.

\begin{landscape}\scriptsize

\begin{longtable}[]{@{}
  >{\raggedright\arraybackslash}p{(\columnwidth - 16\tabcolsep) * \real{0.0938}}
  >{\raggedright\arraybackslash}p{(\columnwidth - 16\tabcolsep) * \real{0.0938}}
  >{\raggedleft\arraybackslash}p{(\columnwidth - 16\tabcolsep) * \real{0.1250}}
  >{\raggedright\arraybackslash}p{(\columnwidth - 16\tabcolsep) * \real{0.0938}}
  >{\raggedleft\arraybackslash}p{(\columnwidth - 16\tabcolsep) * \real{0.1250}}
  >{\raggedleft\arraybackslash}p{(\columnwidth - 16\tabcolsep) * \real{0.1250}}
  >{\raggedleft\arraybackslash}p{(\columnwidth - 16\tabcolsep) * \real{0.1250}}
  >{\raggedright\arraybackslash}p{(\columnwidth - 16\tabcolsep) * \real{0.0938}}
  >{\raggedleft\arraybackslash}p{(\columnwidth - 16\tabcolsep) * \real{0.1250}}@{}}
\toprule\noalign{}
\begin{minipage}[b]{\linewidth}\raggedright
Pair
\end{minipage} & \begin{minipage}[b]{\linewidth}\raggedright
Sign
\end{minipage} & \begin{minipage}[b]{\linewidth}\raggedleft
Rank
\end{minipage} & \begin{minipage}[b]{\linewidth}\raggedright
ROI
\end{minipage} & \begin{minipage}[b]{\linewidth}\raggedleft
\(\rho_A\)
\end{minipage} & \begin{minipage}[b]{\linewidth}\raggedleft
\(\rho_B\)
\end{minipage} & \begin{minipage}[b]{\linewidth}\raggedleft
\(\Delta\rho\)
\end{minipage} & \begin{minipage}[b]{\linewidth}\raggedright
95\% CI
\end{minipage} & \begin{minipage}[b]{\linewidth}\raggedleft
\(\min(p_A, p_B)\)
\end{minipage} \\
\midrule\noalign{}
\endhead
\bottomrule\noalign{}
\endlastfoot
Phonemic minus Semantic \emph{(primary)} & positive & 1 & PoCG\_L* &
\(+0.253\) & \(-0.038\) & \(+0.291\) & \([+0.112, +0.460]\) &
\(1.9 \times 10^{-4}\) \\
Phonemic minus Semantic \emph{(primary)} & positive & 2 & PrCG\_L* &
\(+0.214\) & \(-0.053\) & \(+0.266\) & \([+0.085, +0.445]\) &
\(1.7 \times 10^{-3}\) \\
Phonemic minus Semantic \emph{(primary)} & positive & 3 & SLF\_L* &
\(+0.281\) & \(+0.054\) & \(+0.227\) & \([+0.062, +0.393]\) &
\(3.2 \times 10^{-5}\) \\
Phonemic minus Semantic \emph{(primary)} & positive & 4 & SMG\_L* &
\(+0.270\) & \(+0.072\) & \(+0.198\) & \([+0.014, +0.379]\) &
\(6.7 \times 10^{-5}\) \\
Phonemic minus Semantic \emph{(primary)} & positive & 5 & RLIC\_L &
\(+0.216\) & \(+0.084\) & \(+0.133\) & \([-0.065, +0.325]\) &
\(1.5 \times 10^{-3}\) \\
Phonemic minus Semantic \emph{(primary)} & positive & 6 & SCR\_L &
\(+0.082\) & \(-0.045\) & \(+0.128\) & \([-0.058, +0.310]\) & 0.231 \\
Phonemic minus Semantic \emph{(primary)} & negative & 1 & Caud\_L &
\(-0.073\) & \(+0.053\) & \(-0.126\) & \([-0.334, +0.078]\) & 0.288 \\
Phonemic minus Semantic \emph{(primary)} & negative & 2 & SCC\_L &
\(+0.016\) & \(+0.141\) & \(-0.124\) & \([-0.304, +0.049]\) & 0.040 \\
Phonemic minus Semantic \emph{(primary)} & negative & 3 & TAP\_L &
\(+0.066\) & \(+0.170\) & \(-0.104\) & \([-0.282, +0.083]\) & 0.013 \\
Phonemic minus Semantic \emph{(primary)} & negative & 4 & ALIC\_L &
\(-0.005\) & \(+0.098\) & \(-0.103\) & \([-0.301, +0.091]\) & 0.156 \\
Phonemic minus Semantic \emph{(primary)} & negative & 5 & MOG\_L &
\(+0.163\) & \(+0.262\) & \(-0.098\) & \([-0.276, +0.091]\) &
\(1.1 \times 10^{-4}\) \\
Phonemic minus Semantic \emph{(primary)} & negative & 6 & Cu\_L &
\(+0.099\) & \(+0.185\) & \(-0.086\) & \([-0.282, +0.111]\) &
\(6.8 \times 10^{-3}\) \\
Phonemic minus Neologism \emph{(cross-check)} & positive & 1 & LG\_L* &
\(+0.143\) & \(-0.017\) & \(+0.160\) & \([+0.033, +0.291]\) & 0.037 \\
Phonemic minus Neologism \emph{(cross-check)} & positive & 2 & PTR\_L &
\(+0.183\) & \(+0.080\) & \(+0.102\) & \([-0.026, +0.239]\) &
\(7.5 \times 10^{-3}\) \\
Phonemic minus Neologism \emph{(cross-check)} & positive & 3 & IOG\_L &
\(+0.159\) & \(+0.062\) & \(+0.097\) & \([-0.039, +0.228]\) & 0.020 \\
Phonemic minus Neologism \emph{(cross-check)} & positive & 4 & MOG\_L &
\(+0.163\) & \(+0.078\) & \(+0.085\) & \([-0.040, +0.217]\) & 0.017 \\
Phonemic minus Neologism \emph{(cross-check)} & positive & 5 & Cu\_L &
\(+0.099\) & \(+0.024\) & \(+0.074\) & \([-0.051, +0.198]\) & 0.151 \\
Phonemic minus Neologism \emph{(cross-check)} & positive & 6 & PSTG\_L &
\(+0.216\) & \(+0.144\) & \(+0.071\) & \([-0.058, +0.197]\) &
\(1.5 \times 10^{-3}\) \\
Phonemic minus Neologism \emph{(cross-check)} & negative & 1 &
LenticularFasc\_L & \(+0.074\) & \(+0.182\) & \(-0.108\) &
\([-0.222, +0.006]\) & \(7.8 \times 10^{-3}\) \\
Phonemic minus Neologism \emph{(cross-check)} & negative & 2 & Amyg\_L &
\(+0.086\) & \(+0.184\) & \(-0.098\) & \([-0.223, +0.033]\) &
\(6.9 \times 10^{-3}\) \\
Phonemic minus Neologism \emph{(cross-check)} & negative & 3 & Mynert\_L
& \(+0.076\) & \(+0.172\) & \(-0.096\) & \([-0.231, +0.037]\) & 0.012 \\
Phonemic minus Neologism \emph{(cross-check)} & negative & 4 &
AnsaLenticularis\_L & \(+0.088\) & \(+0.180\) & \(-0.093\) &
\([-0.222, +0.033]\) & \(8.4 \times 10^{-3}\) \\
Phonemic minus Neologism \emph{(cross-check)} & negative & 5 & Put\_L &
\(+0.072\) & \(+0.150\) & \(-0.077\) & \([-0.213, +0.052]\) & 0.029 \\
Phonemic minus Neologism \emph{(cross-check)} & negative & 6 & Hippo\_L
& \(+0.030\) & \(+0.104\) & \(-0.074\) & \([-0.199, +0.048]\) & 0.130 \\
Semantic minus Neologism \emph{(cross-check)} & positive & 1 & LG\_L &
\(+0.204\) & \(-0.017\) & \(+0.221\) & \([-0.001, +0.409]\) &
\(2.8 \times 10^{-3}\) \\
Semantic minus Neologism \emph{(cross-check)} & positive & 2 & MOG\_L &
\(+0.262\) & \(+0.078\) & \(+0.183\) & \([-0.019, +0.376]\) &
\(1.1 \times 10^{-4}\) \\
Semantic minus Neologism \emph{(cross-check)} & positive & 3 & IOG\_L &
\(+0.230\) & \(+0.062\) & \(+0.168\) & \([-0.032, +0.373]\) &
\(7.3 \times 10^{-4}\) \\
Semantic minus Neologism \emph{(cross-check)} & positive & 4 & Cu\_L &
\(+0.185\) & \(+0.024\) & \(+0.160\) & \([-0.039, +0.355]\) &
\(6.8 \times 10^{-3}\) \\
Semantic minus Neologism \emph{(cross-check)} & positive & 5 & PTR\_L &
\(+0.225\) & \(+0.080\) & \(+0.145\) & \([-0.045, +0.343]\) &
\(9.2 \times 10^{-4}\) \\
Semantic minus Neologism \emph{(cross-check)} & positive & 6 & SCC\_L &
\(+0.141\) & \(+0.011\) & \(+0.130\) & \([-0.059, +0.303]\) & 0.040 \\
Semantic minus Neologism \emph{(cross-check)} & negative & 1 & PoCG\_L*
& \(-0.038\) & \(+0.248\) & \(-0.286\) & \([-0.469, -0.104]\) &
\(2.6 \times 10^{-4}\) \\
Semantic minus Neologism \emph{(cross-check)} & negative & 2 & PrCG\_L*
& \(-0.053\) & \(+0.196\) & \(-0.248\) & \([-0.451, -0.060]\) &
\(4.2 \times 10^{-3}\) \\
Semantic minus Neologism \emph{(cross-check)} & negative & 3 & SLF\_L* &
\(+0.054\) & \(+0.240\) & \(-0.186\) & \([-0.360, -0.004]\) &
\(4.2 \times 10^{-4}\) \\
Semantic minus Neologism \emph{(cross-check)} & negative & 4 & Amyg\_L &
\(+0.013\) & \(+0.184\) & \(-0.171\) & \([-0.398, +0.051]\) &
\(6.9 \times 10^{-3}\) \\
Semantic minus Neologism \emph{(cross-check)} & negative & 5 & SMG\_L &
\(+0.072\) & \(+0.234\) & \(-0.162\) & \([-0.349, +0.032]\) &
\(5.8 \times 10^{-4}\) \\
Semantic minus Neologism \emph{(cross-check)} & negative & 6 &
LenticularFasc\_L & \(+0.021\) & \(+0.182\) & \(-0.161\) &
\([-0.335, +0.006]\) & \(7.8 \times 10^{-3}\) \\
\end{longtable}

\end{landscape}

The primary Phonemic \textgreater{} Semantic direction is anchored by
four ROIs whose bootstrap 95\% CI excludes zero: PoCG\_L, PrCG\_L,
SLF\_L, SMG\_L, the frontal-parietal / perisylvian territory that human
lesion-symptom mapping has historically associated with
phonological-output deficits. The same territory is recovered by the
converse direction of the Semantic-vs-Neologism cross-check (Neologism
\textgreater{} Semantic; PoCG\_L, PrCG\_L, SLF\_L all CI-significant),
giving independent corroboration of the cluster from a contrast against
a different baseline. The Semantic \textgreater{} Phonemic direction of
the primary contrast has consistently signed point estimates
(\(\Delta\rho\) between \(-0.10\) and \(-0.13\)) in left posterior
temporo-occipital cortex (MOG\_L, Cu\_L, IOG\_L) and subcortical
structures (Caud\_L, TAP\_L, ALIC\_L, SCC\_L), but the bootstrap CIs at
\(N = 213\) all cross zero, reflecting the broader spatial distribution
and smaller per-region magnitude of the semantic-favoring axis (the same
asymmetry seen on the LLM side, where semantic-favoring effects are
roughly an order of magnitude weaker than phonemic-favoring effects).
The Semantic \textgreater{} Neologism cross-check picks up the same
posterior territory (LG\_L, MOG\_L, IOG\_L, Cu\_L), with one ROI (LG\_L)
narrowly excluding zero.

Machine-readable CSVs: per-(error, ROI) Spearman correlations in
\texttt{code/tfce\_results/lsm\_roi/lsm\_roi\_correlations.csv};
per-(pair, ROI) correlation differences with bootstrap CIs in
\texttt{code/tfce\_results/lsm\_roi/lsm\_roi\_subtraction.csv}.

\textbf{Table S3.} Human cortex ROI inference across all 15 pairwise
error-category contrasts

Per-pair summary of the univariate Spearman correlation-difference VLSM
run on the \(N = 213\) patient cohort across all 15 ordered pairs from
the six non-Correct PNT error categories (Semantic, Unrelated, Phonemic,
Mixed, Neologism, NoResponse). Pair naming follows the LLM all-pairs
convention \texttt{A\_minus\_B} from main-text Figure 3 and
Supplementary Figure S2. For each pair we give: the two directions, the
count of left-hemisphere JHU ROIs (of 64) with patient-level bootstrap
95\% CI excluding zero in that direction (\#sig), the most-extreme
correlation-difference \(\Delta\rho\) observed in that direction, and
the top three ROIs by \(|\Delta\rho|\) (starred when CI excludes zero).

\begin{landscape}\scriptsize

\begin{longtable}[]{@{}
  >{\raggedright\arraybackslash}p{(\columnwidth - 16\tabcolsep) * \real{0.0968}}
  >{\raggedright\arraybackslash}p{(\columnwidth - 16\tabcolsep) * \real{0.0968}}
  >{\raggedleft\arraybackslash}p{(\columnwidth - 16\tabcolsep) * \real{0.1290}}
  >{\raggedleft\arraybackslash}p{(\columnwidth - 16\tabcolsep) * \real{0.1290}}
  >{\raggedright\arraybackslash}p{(\columnwidth - 16\tabcolsep) * \real{0.0968}}
  >{\raggedright\arraybackslash}p{(\columnwidth - 16\tabcolsep) * \real{0.0968}}
  >{\raggedleft\arraybackslash}p{(\columnwidth - 16\tabcolsep) * \real{0.1290}}
  >{\raggedleft\arraybackslash}p{(\columnwidth - 16\tabcolsep) * \real{0.1290}}
  >{\raggedright\arraybackslash}p{(\columnwidth - 16\tabcolsep) * \real{0.0968}}@{}}
\toprule\noalign{}
\begin{minipage}[b]{\linewidth}\raggedright
Pair (A minus B)
\end{minipage} & \begin{minipage}[b]{\linewidth}\raggedright
A \textgreater{} B direction
\end{minipage} & \begin{minipage}[b]{\linewidth}\raggedleft
\#sig
\end{minipage} & \begin{minipage}[b]{\linewidth}\raggedleft
max \(\Delta\rho\)
\end{minipage} & \begin{minipage}[b]{\linewidth}\raggedright
top 3 ROIs
\end{minipage} & \begin{minipage}[b]{\linewidth}\raggedright
B \textgreater{} A direction
\end{minipage} & \begin{minipage}[b]{\linewidth}\raggedleft
\#sig
\end{minipage} & \begin{minipage}[b]{\linewidth}\raggedleft
min \(\Delta\rho\)
\end{minipage} & \begin{minipage}[b]{\linewidth}\raggedright
top 3 ROIs
\end{minipage} \\
\midrule\noalign{}
\endhead
\bottomrule\noalign{}
\endlastfoot
Semantic minus Unrelated & Semantic \textgreater{} Unrelated & 0 & +0.00
& & Unrelated \textgreater{} Semantic & 31 & \(-0.28\) & SLF\_L\emph{,
PrCG\_L}, PoCG\_L* \\
Semantic minus Phonemic \emph{(primary)} & Semantic \textgreater{}
Phonemic & 0 & \(+0.13\) & Caud\_L, SCC\_L, TAP\_L & Phonemic
\textgreater{} Semantic & 4 & \(-0.29\) & PoCG\_L\emph{, PrCG\_L},
SLF\_L* \\
Semantic minus Mixed & Semantic \textgreater{} Mixed & 4 & \(+0.23\) &
MOG\_L\emph{, SOG\_L}, Cu\_L* & Mixed \textgreater{} Semantic & 1 &
\(-0.16\) & LenticularFasc\_L*, CP\_L, AnsaLenticularis\_L \\
Semantic minus Neologism \emph{(cross-check)} & Semantic \textgreater{}
Neologism & 1 & \(+0.22\) & LG\_L*, MOG\_L, IOG\_L & Neologism
\textgreater{} Semantic & 3 & \(-0.29\) & PoCG\_L\emph{, PrCG\_L},
SLF\_L* \\
Semantic minus NoResponse & Semantic \textgreater{} NoResponse & 0 &
+0.00 & & NoResponse \textgreater{} Semantic & 47 & \(-0.40\) &
SLF\_L\emph{, EC\_L}, Fx/ST\_L* \\
Unrelated minus Phonemic & Unrelated \textgreater{} Phonemic & 24 &
\(+0.27\) & Hippo\_L\emph{, SS\_L}, Caud\_L* & Phonemic \textgreater{}
Unrelated & 0 & \(-0.02\) & PoCG\_L \\
Unrelated minus Mixed & Unrelated \textgreater{} Mixed & 45 & \(+0.31\)
& SS\_L\emph{, SLF\_L}, IFG\_orbitalis\_L* & Mixed \textgreater{}
Unrelated & 0 & \(-0.04\) & LenticularFasc\_L \\
Unrelated minus Neologism & Unrelated \textgreater{} Neologism & 27 &
\(+0.27\) & LG\_L\emph{, FuG\_L}, IOG\_L* & Neologism \textgreater{}
Unrelated & 0 & \(-0.04\) & LenticularFasc\_L, PoCG\_L, OpticTract\_L \\
Unrelated minus NoResponse & Unrelated \textgreater{} NoResponse & 0 &
\(+0.06\) & CGC\_L, Thal\_L, SFG\_L & NoResponse \textgreater{}
Unrelated & 1 & \(-0.12\) & STG\_L, SLF\_L*, PSIG\_L \\
Phonemic minus Mixed & Phonemic \textgreater{} Mixed & 7 & \(+0.25\) &
SLF\_L\emph{, PoCG\_L}, SMG\_L* & Mixed \textgreater{} Phonemic & 0 &
\(-0.11\) & LenticularFasc\_L, GP\_L, SCC\_L \\
Phonemic minus Neologism \emph{(cross-check)} & Phonemic \textgreater{}
Neologism & 1 & \(+0.16\) & LG\_L*, PTR\_L, IOG\_L & Neologism
\textgreater{} Phonemic & 0 & \(-0.11\) & LenticularFasc\_L, Amyg\_L,
Mynert\_L \\
Phonemic minus NoResponse & Phonemic \textgreater{} NoResponse & 0 &
+0.00 & & NoResponse \textgreater{} Phonemic & 38 & \(-0.33\) &
PSIG\_L\emph{, Hippo\_L}, SS\_L* \\
Mixed minus Neologism & Mixed \textgreater{} Neologism & 0 & \(+0.09\) &
SCC\_L, LG\_L, TAP\_L & Neologism \textgreater{} Mixed & 8 & \(-0.23\) &
PoCG\_L\emph{, MFG\_L}, SLF\_L* \\
Mixed minus NoResponse & Mixed \textgreater{} NoResponse & 0 & \(+0.02\)
& LenticularFasc\_L & NoResponse \textgreater{} Mixed & 47 & \(-0.42\) &
SLF\_L\emph{, PSTG\_L}, PTR\_L* \\
Neologism minus NoResponse & Neologism \textgreater{} NoResponse & 0 &
\(+0.02\) & LenticularFasc\_L & NoResponse \textgreater{} Neologism & 35
& \(-0.37\) & PSIG\_L\emph{, PTR\_L}, MOG\_L* \\
\end{longtable}

\end{landscape}

The cross-substrate pattern is most striking for the contrasts that
involve a sharply specialized real-word error class against a
non-specific baseline. Phonemic and Unrelated each pull the
frontal-perisylvian PoCG/PrCG/SLF cluster apart from any less-targeted
contrast partner; Semantic and Neologism each pull the posterior
temporo-occipital MOG/IOG/Cu cluster apart from less-targeted partners.
The NoResponse and Unrelated rows show that error-rate-dominated
contrasts (where one category is essentially driven by total impairment)
load broadly across many ROIs, consistent with the absence of
category-specific spatial selectivity that the primary
Semantic-vs-Phonemic contrast tests for.

Machine-readable CSV:
\texttt{code/tfce\_results/lsm\_roi/lsm\_roi\_subtraction.csv} (one row
per pair × ROI; 960 rows total).

\textbf{Table S4.} Per-cluster dose-response slopes across all 15
pairwise contrasts

OLS dose-response slopes \(\hat\beta_\sigma\) and \(\hat\beta_\rho\) on
the per-cell group means \(\bar D(\sigma, \rho \mid L^\star)\) at each
surviving cluster's layers, with 1,000-iterate cluster-bootstrap 95\%
CIs (over seeds). Discovery (40 seeds) and validation (40 seeds)
reported side by side. ``n/a'' means no surviving cluster in that
pair-direction on that half. Per Julius's round-3 instruction, slopes
are reported as effect-size estimates with bootstrap intervals rather
than as p-values, the latter are trivially extreme at this sample size
by the central-limit argument and would obscure rather than inform.

\begin{landscape}\scriptsize

\begin{longtable}[]{@{}
  >{\raggedright\arraybackslash}p{(\columnwidth - 14\tabcolsep) * \real{0.1250}}
  >{\raggedright\arraybackslash}p{(\columnwidth - 14\tabcolsep) * \real{0.1250}}
  >{\raggedright\arraybackslash}p{(\columnwidth - 14\tabcolsep) * \real{0.1250}}
  >{\raggedright\arraybackslash}p{(\columnwidth - 14\tabcolsep) * \real{0.1250}}
  >{\raggedright\arraybackslash}p{(\columnwidth - 14\tabcolsep) * \real{0.1250}}
  >{\raggedright\arraybackslash}p{(\columnwidth - 14\tabcolsep) * \real{0.1250}}
  >{\raggedright\arraybackslash}p{(\columnwidth - 14\tabcolsep) * \real{0.1250}}
  >{\raggedright\arraybackslash}p{(\columnwidth - 14\tabcolsep) * \real{0.1250}}@{}}
\toprule\noalign{}
\begin{minipage}[b]{\linewidth}\raggedright
Pair (A minus B)
\end{minipage} & \begin{minipage}[b]{\linewidth}\raggedright
Sign
\end{minipage} & \begin{minipage}[b]{\linewidth}\raggedright
Disc layers
\end{minipage} & \begin{minipage}[b]{\linewidth}\raggedright
Val layers
\end{minipage} & \begin{minipage}[b]{\linewidth}\raggedright
\(\hat\beta_\sigma\) disc {[}95\% CI{]}
\end{minipage} & \begin{minipage}[b]{\linewidth}\raggedright
\(\hat\beta_\sigma\) val {[}95\% CI{]}
\end{minipage} & \begin{minipage}[b]{\linewidth}\raggedright
\(\hat\beta_\rho\) disc {[}95\% CI{]}
\end{minipage} & \begin{minipage}[b]{\linewidth}\raggedright
\(\hat\beta_\rho\) val {[}95\% CI{]}
\end{minipage} \\
\midrule\noalign{}
\endhead
\bottomrule\noalign{}
\endlastfoot
Semantic minus Unrelated & negative & 19-20 & 18-20 & -0.127 {[}-0.137,
-0.118{]} & -0.131 {[}-0.141, -0.120{]} & -0.375 {[}-0.400, -0.351{]} &
-0.403 {[}-0.424, -0.378{]} \\
Semantic minus Phonemic \emph{(primary)} & negative & 22-31 & 23-33 &
-0.182 {[}-0.185, -0.179{]} & -0.176 {[}-0.180, -0.171{]} & -0.462
{[}-0.474, -0.451{]} & -0.434 {[}-0.444, -0.423{]} \\
Semantic minus Mixed & positive & 27-29 & 26-29 & +0.022 {[}+0.020,
+0.023{]} & +0.022 {[}+0.021, +0.024{]} & +0.089 {[}+0.084, +0.094{]} &
+0.091 {[}+0.087, +0.095{]} \\
Semantic minus Neologism \emph{(cross-check)} & negative & 19-20 & 18-19
& -0.178 {[}-0.190, -0.167{]} & -0.173 {[}-0.182, -0.164{]} & -0.489
{[}-0.522, -0.458{]} & -0.477 {[}-0.504, -0.449{]} \\
Semantic minus NoResponse & negative & 0-38 & 0-38 & -0.329 {[}-0.339,
-0.319{]} & -0.330 {[}-0.342, -0.319{]} & +0.238 {[}+0.213, +0.261{]} &
+0.232 {[}+0.200, +0.265{]} \\
Unrelated minus Phonemic & negative & 24-34 & 24-37 & -0.170 {[}-0.174,
-0.167{]} & -0.155 {[}-0.158, -0.152{]} & -0.433 {[}-0.443, -0.423{]} &
-0.384 {[}-0.393, -0.377{]} \\
Unrelated minus Phonemic & positive & 8-9 & 6-7 & +0.160 {[}+0.147,
+0.176{]} & +0.122 {[}+0.110, +0.136{]} & +0.448 {[}+0.409, +0.496{]} &
+0.235 {[}+0.202, +0.274{]} \\
Unrelated minus Mixed & positive & 19-22 & 18-20 & +0.116 {[}+0.110,
+0.122{]} & +0.145 {[}+0.133, +0.156{]} & +0.366 {[}+0.353, +0.381{]} &
+0.478 {[}+0.451, +0.503{]} \\
Unrelated minus Neologism & negative & 33-34 & n/a & -0.025 {[}-0.028,
-0.023{]} & n/a & -0.026 {[}-0.034, -0.018{]} & n/a \\
Unrelated minus NoResponse & negative & 0-38 & 0-38 & -0.252 {[}-0.262,
-0.242{]} & -0.254 {[}-0.266, -0.243{]} & +0.429 {[}+0.402, +0.456{]} &
+0.421 {[}+0.390, +0.452{]} \\
Phonemic minus Mixed & positive & 19-36 & 15-38 & +0.175 {[}+0.172,
+0.178{]} & +0.161 {[}+0.158, +0.164{]} & +0.472 {[}+0.464, +0.480{]} &
+0.444 {[}+0.437, +0.451{]} \\
Phonemic minus Neologism \emph{(cross-check)} & negative & n/a & 11-12 &
n/a & -0.063 {[}-0.071, -0.055{]} & n/a & -0.112 {[}-0.136, -0.089{]} \\
Phonemic minus Neologism \emph{(cross-check)} & positive & 24-31 & 24-33
& +0.166 {[}+0.163, +0.169{]} & +0.158 {[}+0.154, +0.161{]} & +0.467
{[}+0.458, +0.474{]} & +0.437 {[}+0.429, +0.445{]} \\
Phonemic minus NoResponse & negative & 0-38 & 0-38 & -0.220 {[}-0.230,
-0.211{]} & -0.222 {[}-0.232, -0.211{]} & +0.511 {[}+0.485, +0.533{]} &
+0.505 {[}+0.474, +0.534{]} \\
Mixed minus Neologism & negative & 19-20 & 31-32 & -0.197 {[}-0.207,
-0.187{]} & -0.035 {[}-0.037, -0.033{]} & -0.582 {[}-0.609, -0.554{]} &
-0.067 {[}-0.072, -0.062{]} \\
Mixed minus NoResponse & negative & 0-38 & 0-38 & -0.340 {[}-0.349,
-0.330{]} & -0.340 {[}-0.352, -0.330{]} & +0.190 {[}+0.166, +0.214{]} &
+0.184 {[}+0.154, +0.215{]} \\
Neologism minus NoResponse & negative & 2-38 & 0-38 & -0.234 {[}-0.245,
-0.224{]} & -0.262 {[}-0.273, -0.251{]} & +0.456 {[}+0.432, +0.480{]} &
+0.353 {[}+0.320, +0.385{]} \\
\end{longtable}

\end{landscape}

Headline reading of Table S4: across the three pre-registered
confirmatory contrasts (rows starred \emph{(primary)} and
\emph{(cross-check)}), the surviving phonemic-favoring direction's
\(\hat\beta_\sigma\) and \(\hat\beta_\rho\) are of the same sign as the
cluster's contrast (deeper / more extensive perturbation → larger
directional contrast) and replicate within \$\sim\$10\% across the seed
split. The remaining (exploratory) rows show that
contrast-vs-error-class slopes scale similarly for most pair-directions,
with the largest dose-response shifts driven by NoResponse-baseline
contrasts where total error-rate effects dominate.

Machine-readable:
\texttt{code/tfce\_results/llm\_dose\_response/disc\_dose\_cluster\_slopes.csv}
and \texttt{val\_dose\_cluster\_slopes.csv}; per-cell surfaces in
\texttt{*\_cluster\_surface.csv}.

\textbf{Table S5.} Cortex per-category univariate top ROIs (Spearman
correlations, \(N = 213\) patients)

Top five LH JHU ROIs per non-Correct PNT error category, ranked by
\(|\rho_{A,r}|\). \(\rho_{A,r}\) is the Spearman rank correlation
between the per-patient \(A\)-category PNT proportion and the
per-patient lesion load in ROI \(r\). \(p\) is the two-sided Spearman
\(p\)-value against an unstructured null. No age, sex, or other
covariate is adjusted for (matched-no-covariates design to align with
the LLM-side analysis in Table S6). Stars indicate
\(p < \alpha'_{64} = 0.05 / 64 \approx 7.8 \times 10^{-4}\)
(per-category Bonferroni across 64 ROIs); double-starred entries
additionally clear
\(\alpha'_{384} = 0.05 / 384 \approx 1.3 \times 10^{-4}\) (panel-wide
Bonferroni across 6 categories \(\times\) 64 ROIs).

\begin{longtable}[]{@{}lrlrr@{}}
\toprule\noalign{}
Category & Rank & ROI & Spearman \(\rho\) & \(p\) \\
\midrule\noalign{}
\endhead
\bottomrule\noalign{}
\endlastfoot
Phonemic & 1 & SLF\_L** & \(+0.281\) & \(3.2 \times 10^{-5}\) \\
Phonemic & 2 & SMG\_L** & \(+0.270\) & \(6.7 \times 10^{-5}\) \\
Phonemic & 3 & PoCG\_L* & \(+0.253\) & \(1.9 \times 10^{-4}\) \\
Phonemic & 4 & RLIC\_L & \(+0.216\) & \(1.5 \times 10^{-3}\) \\
Phonemic & 5 & PSTG\_L & \(+0.216\) & \(1.5 \times 10^{-3}\) \\
Semantic & 1 & MOG\_L** & \(+0.262\) & \(1.1 \times 10^{-4}\) \\
Semantic & 2 & IOG\_L* & \(+0.230\) & \(7.3 \times 10^{-4}\) \\
Semantic & 3 & PTR\_L & \(+0.225\) & \(9.2 \times 10^{-4}\) \\
Semantic & 4 & PSMG\_L & \(+0.224\) & \(1.0 \times 10^{-3}\) \\
Semantic & 5 & PSTG\_L & \(+0.210\) & \(2.1 \times 10^{-3}\) \\
Neologism & 1 & PoCG\_L* & \(+0.248\) & \(2.6 \times 10^{-4}\) \\
Neologism & 2 & SLF\_L* & \(+0.240\) & \(4.2 \times 10^{-4}\) \\
Neologism & 3 & SMG\_L* & \(+0.234\) & \(5.8 \times 10^{-4}\) \\
Neologism & 4 & PrCG\_L & \(+0.196\) & \(4.2 \times 10^{-3}\) \\
Neologism & 5 & IFO\_L & \(+0.190\) & \(5.5 \times 10^{-3}\) \\
Mixed & 1 & LenticularFasc\_L & \(+0.182\) & \(7.9 \times 10^{-3}\) \\
Mixed & 2 & GP\_L & \(+0.166\) & \(1.6 \times 10^{-2}\) \\
Mixed & 3 & AnsaLenticularis\_L & \(+0.165\) & \(1.6 \times 10^{-2}\) \\
Mixed & 4 & MTG\_L & \(+0.122\) & \(7.5 \times 10^{-2}\) \\
Mixed & 5 & UNC\_L & \(+0.111\) & \(1.1 \times 10^{-1}\) \\
Unrelated & 1 & SS\_L** & \(+0.374\) & \(1.8 \times 10^{-8}\) \\
Unrelated & 2 & RLIC\_L** & \(+0.347\) & \(2.0 \times 10^{-7}\) \\
Unrelated & 3 & PSTG\_L** & \(+0.336\) & \(4.9 \times 10^{-7}\) \\
Unrelated & 4 & FuG\_L** & \(+0.334\) & \(6.1 \times 10^{-7}\) \\
Unrelated & 5 & AG\_L** & \(+0.333\) & \(6.4 \times 10^{-7}\) \\
NoResponse & 1 & SLF\_L** & \(+0.452\) & \(< 10^{-10}\) \\
NoResponse & 2 & PSTG\_L** & \(+0.445\) & \(< 10^{-10}\) \\
NoResponse & 3 & STG\_L** & \(+0.444\) & \(< 10^{-10}\) \\
NoResponse & 4 & PTR\_L** & \(+0.438\) & \(< 10^{-10}\) \\
NoResponse & 5 & MOG\_L** & \(+0.425\) & \(< 10^{-10}\) \\
\end{longtable}

Headline reading of Table S5: every non-Correct PNT category carries
category-specific univariate signal on the cortex before any subtraction
is applied. Phonemic, Semantic, Neologism, Unrelated, and NoResponse
each have at least one ROI clearing the panel-wide Bonferroni threshold;
Mixed is the only category whose top ROI does not clear even the
per-category Bonferroni, consistent with Mixed being a heterogeneous
semantic-phonemic blend that lacks a dedicated cortical correlate. The
Phonemic and Neologism top-ROI lists overlap heavily (SLF\_L, SMG\_L,
PoCG\_L on both), and the Semantic top-ROI list shares perisylvian ROIs
(PSTG\_L) with Phonemic, exactly the shared loading that the Phonemic vs
Semantic subtraction in main-text Figures 3 and 4 removes to isolate the
differential frontal-perisylvian residual.

Machine-readable:
\texttt{code/tfce\_results/lsm\_roi/lsm\_roi\_correlations.csv}
(per-(error, ROI) Spearman correlations on the full 213-patient cohort).

\textbf{Table S6.} LLM per-category univariate peak-layer summary (40
discovery seeds)

Per-category peak-layer summary on the LLM side, computed from the
per-(seed, layer) mean PNT proportion \(\bar p_{A, s}(\ell)\) with
\(\sigma\) and \(\rho\) averaged within each seed (§S7.2). The group
statistic is the across-seed mean \(\bar p_A(\ell)\) on the 40 discovery
seeds. ``Peak layer'' is the \(\ell\) at which \(\bar p_A(\ell)\) is
maximal; ``peak proportion'' is \(\bar p_A\) at that peak with 95\% CI
across seeds; ``half-max span'' is the contiguous range of layers at
which \(\bar p_A(\ell) \geq 0.5 \cdot \max_\ell \bar p_A(\ell)\),
indexing the width of the per-category layer profile.

\begin{longtable}[]{@{}
  >{\raggedright\arraybackslash}p{(\columnwidth - 8\tabcolsep) * \real{0.1765}}
  >{\raggedleft\arraybackslash}p{(\columnwidth - 8\tabcolsep) * \real{0.2353}}
  >{\raggedright\arraybackslash}p{(\columnwidth - 8\tabcolsep) * \real{0.1765}}
  >{\raggedright\arraybackslash}p{(\columnwidth - 8\tabcolsep) * \real{0.1765}}
  >{\raggedleft\arraybackslash}p{(\columnwidth - 8\tabcolsep) * \real{0.2353}}@{}}
\toprule\noalign{}
\begin{minipage}[b]{\linewidth}\raggedright
Category
\end{minipage} & \begin{minipage}[b]{\linewidth}\raggedleft
Peak layer
\end{minipage} & \begin{minipage}[b]{\linewidth}\raggedright
Peak proportion {[}95\% CI{]}
\end{minipage} & \begin{minipage}[b]{\linewidth}\raggedright
Half-max span (layers)
\end{minipage} & \begin{minipage}[b]{\linewidth}\raggedleft
Grand-mean proportion
\end{minipage} \\
\midrule\noalign{}
\endhead
\bottomrule\noalign{}
\endlastfoot
Phonemic & 29 & \(0.069\) \([0.068, 0.071]\) & 14--39 (26 layers) &
\(0.042\) \\
Semantic & 39 & \(0.025\) \([0.024, 0.026]\) & 17--39 (23 layers) &
\(0.012\) \\
Neologism & 16 & \(0.063\) \([0.061, 0.066]\) & 1--22 (22 layers) &
\(0.034\) \\
Mixed & 39 & \(0.010\) \([0.009, 0.010]\) & 15--39 (25 layers) &
\(0.005\) \\
Unrelated & 9 & \(0.089\) \([0.085, 0.093]\) & 4--20 (17 layers) &
\(0.040\) \\
NoResponse & 0 & \(0.719\) \([0.700, 0.738]\) & 0--22 (23 layers) &
\(0.417\) \\
\end{longtable}

Headline reading of Table S6: every non-Correct PNT category has a
coherent peaked layer profile on the LLM. Neologism peaks early-mid
network (L16) with the narrowest profile; Unrelated peaks earlier (L9);
Phonemic peaks late-mid (L29) with a broad late tail; Semantic peaks at
the last layer (L39) but has the smallest peak proportion. NoResponse is
dominated by the L0 cell (unperturbed-layer noise injection drives the
model to refuse), an unsurprising boundary effect of the perturbation
grid; we report it for completeness. The Phonemic and Semantic peak
layers (29 vs 39) and their broad late half-max spans (L14--39 vs
L17--39) overlap substantially, paralleling the shared cortical loading
documented in Table S5, and again, the Phonemic vs Semantic subtraction
along the layer axis (main-text Figures 3--4) isolates the
\emph{differential} portion of that overlap (layers 22--31).

Machine-readable: recomputed at request time from
\texttt{PNT\_80\_seeds\_discovery\_40\_summary.csv} via the per-category
aggregation in
\texttt{code/plot\_figure2\_per\_category\_univariate\_v1.py}.

\textbf{Supplementary Figures}

\includegraphics[width=1\textwidth,height=\textheight]{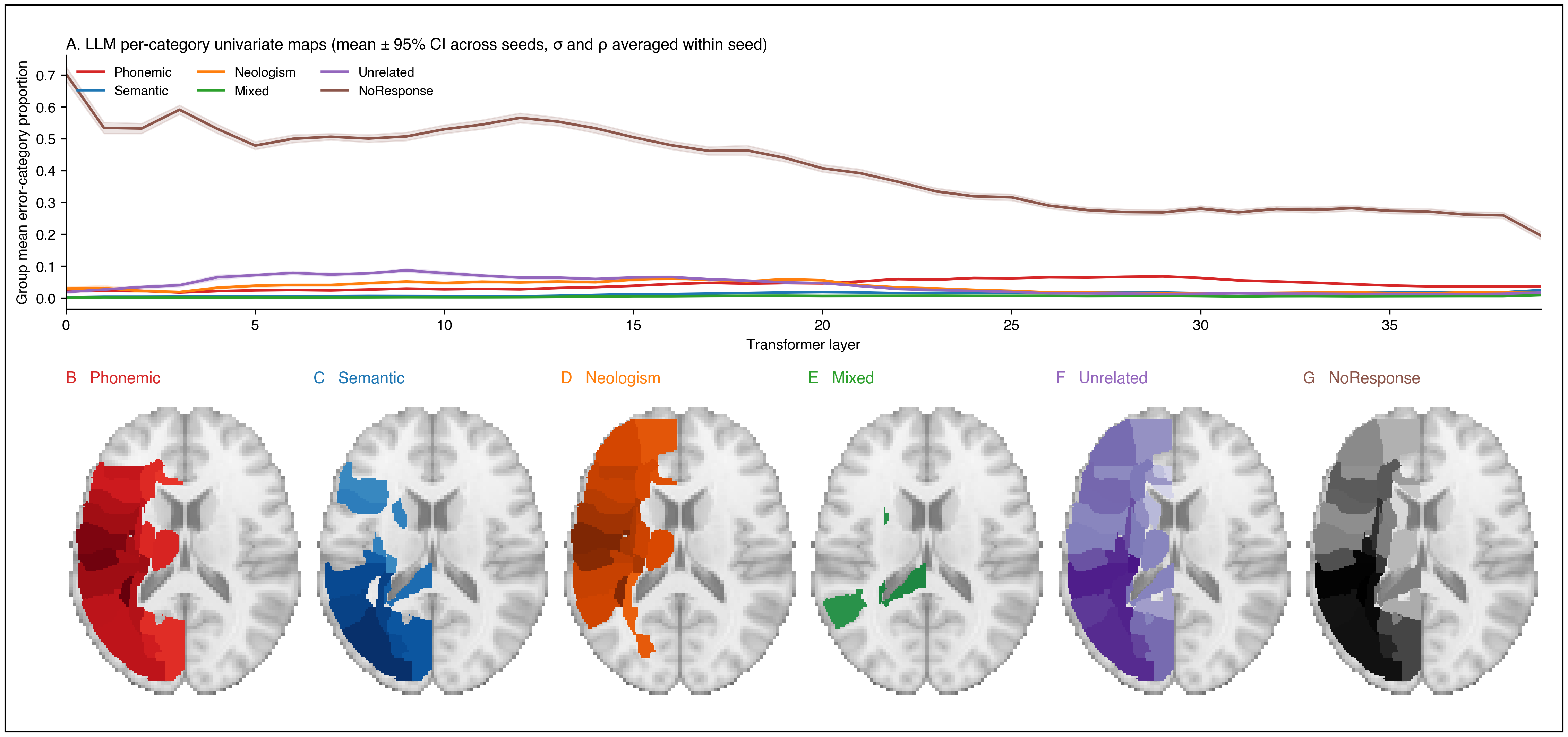}

\textbf{Figure S1.} Per-category univariate maps on both substrates

LLM-side per-category univariate maps (top row): the per-layer group
mean (and 95\% CI across seeds) of each PNT error category's proportion
under perturbation, with \(\sigma\) and \(\rho\) averaged within seed.
Cortex-side per-category univariate maps (bottom row, B--G): the per-ROI
Spearman rank correlation between per-patient error-category proportion
and per-patient JHU ROI lesion load, rendered on the MNI152 T1 template
(axial slice \(z = 16\)), with one panel per category colored by
category. The figure shows the marginals that the main-text subtraction
figures are built on: notably, the Phonemic and Semantic univariate maps
both load onto overlapping perisylvian territory, so their subtraction
(Figure 2) reflects the \emph{differential} loading rather than two
clearly separable spatial signals, explaining why the semantic-favoring
subtraction direction is sub-threshold without that being a failure of
the method.

\includegraphics[width=1\textwidth,height=\textheight]{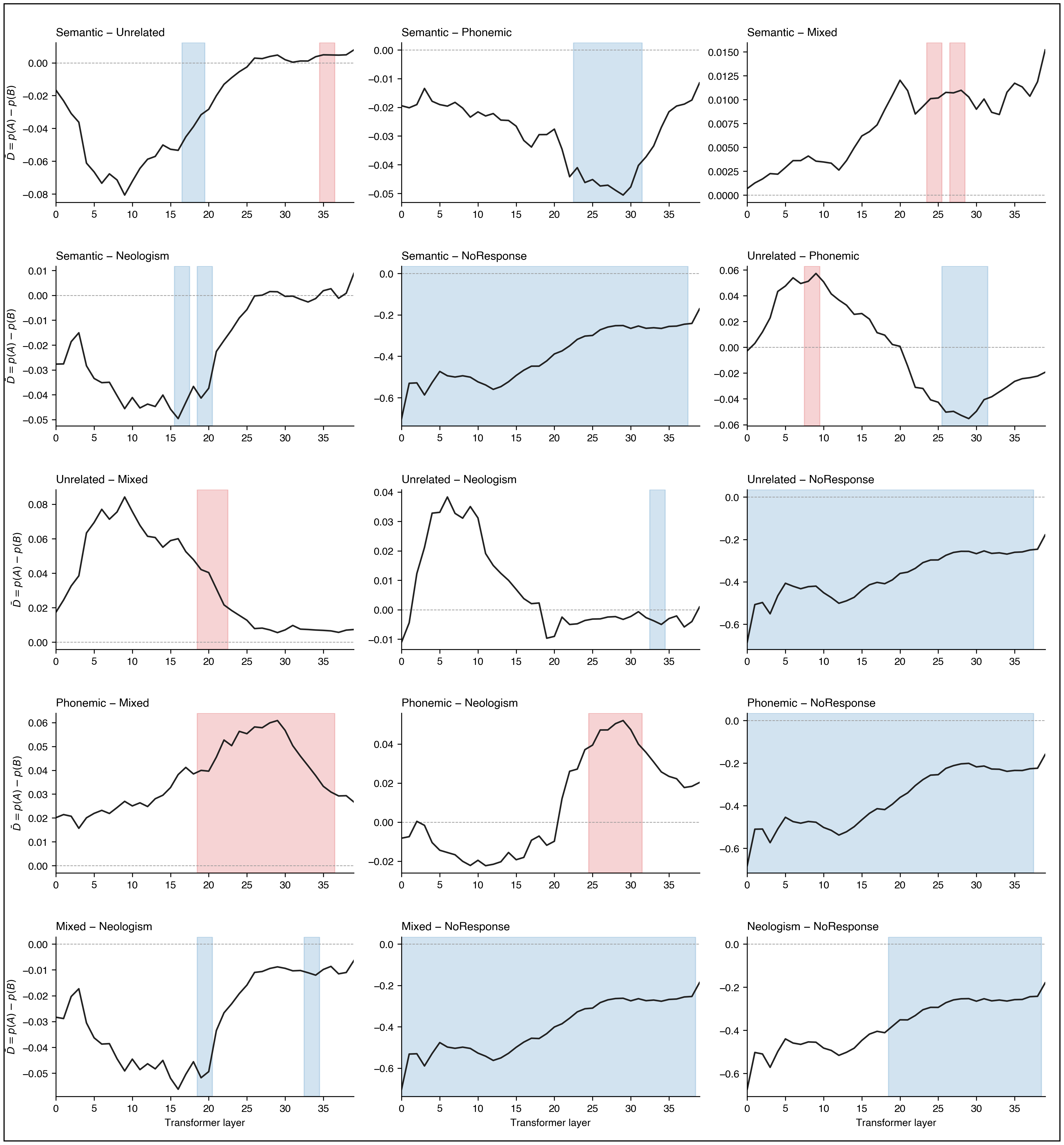}

\textbf{Figure S2.} LLM: per-pair layer-axis maps for all 15 pairwise
subtraction contrasts

A multi-panel figure showing, for each of the 15 ordered pairs across
the six non-Correct PNT error categories, the group-level layer-axis
profile \(\bar D(\ell)\) (40 discovery seeds, \(\sigma\) and \(\rho\)
averaged within seed). TFCE-surviving layer ranges are shaded by
direction (red = positive, \(A > B\); blue = negative, \(B > A\)). This
is the supplementary all-pairs companion to main-text Figure 3.
Per-cluster statistics for all 15 contrasts are in machine-readable form
at
\texttt{code/tfce\_results/llm\_seed\_subject/disc\_llm\_region\_summary.csv}
(discovery) and
\texttt{code/tfce\_results/llm\_seed\_subject/val\_llm\_region\_summary.csv}
(validation).

\includegraphics[width=1\textwidth,height=\textheight]{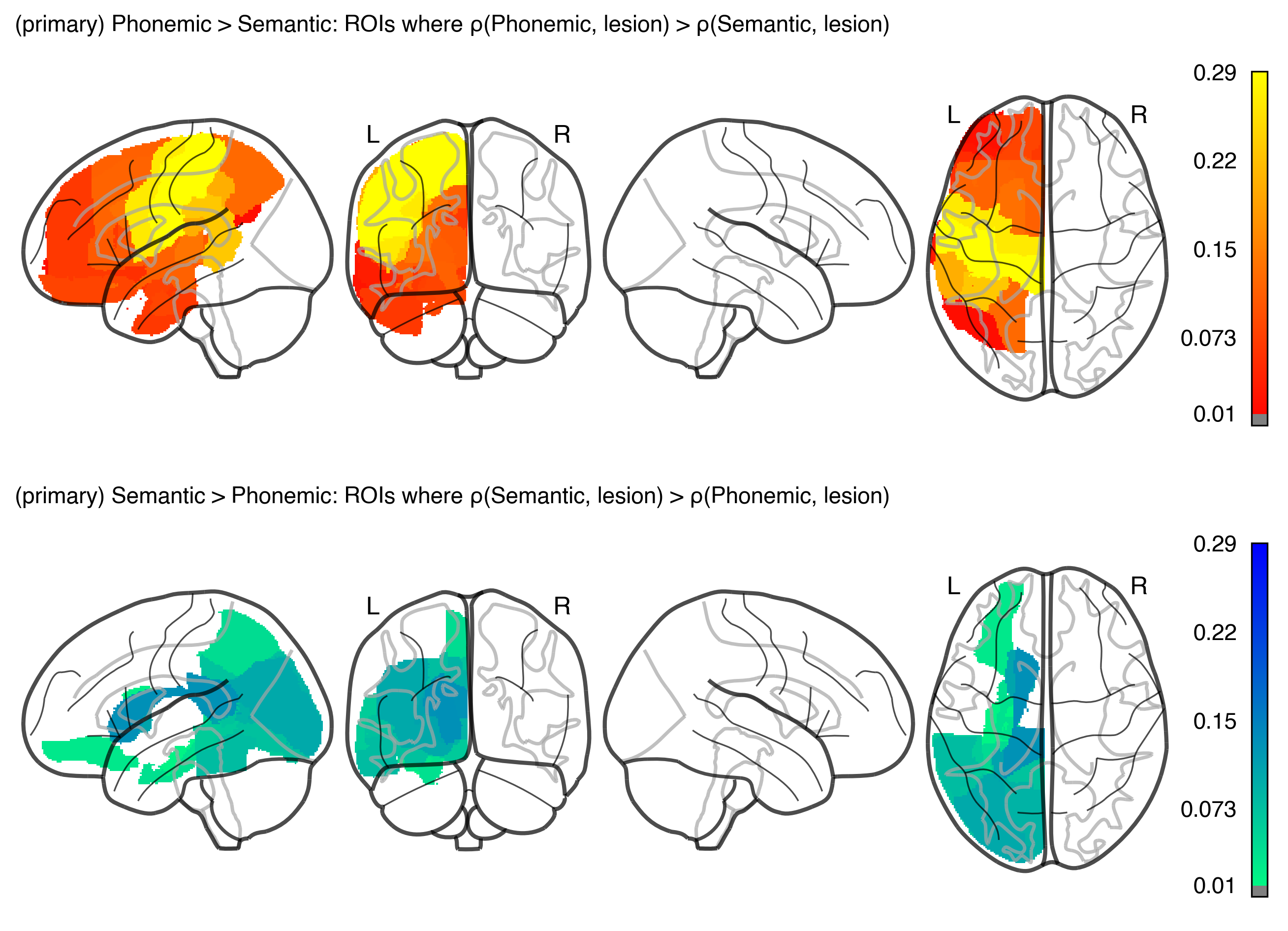}

\textbf{Figure S3.} Human cortex: per-pair ROI correlation-difference
maps for all 15 pairwise subtraction contrasts

A multi-page PDF in which every page renders one pair (top panel
\(A > B\), bottom panel \(B > A\)) for the full set of 15 ordered
pairwise contrasts across the six non-Correct PNT error categories. Each
panel shows the 64 left-hemisphere JHU atlas ROIs colored by per-ROI
correlation difference
\(\Delta\rho_r = \rho(A, \text{lesion}_r) - \rho(B, \text{lesion}_r)\),
where \(\rho\) is the Spearman rank correlation between the per-patient
error-type proportion and the per-patient ROI lesion load on the
\(N = 213\) patient cohort. Warm colormap encodes \(A > B\) (positive
\(\Delta\rho\)); cool colormap encodes \(B > A\) (negative
\(\Delta\rho\)). Pair naming and ordering match the LLM all-pairs
convention in Figure S2, allowing direct visual comparison between
substrates pair by pair. Per-pair summary statistics (count of ROIs with
bootstrap 95\% CI excluding zero, top three ROIs in each direction) are
in Table S3.

Multi-page PDF:
\texttt{figures/lsm\_roi/supplementary\_brain\_roi\_maps.pdf}. A single
thumbnail panel for the primary Phonemic-vs-Semantic contrast is
embedded here for reference:

\textbf{Figure S4.} Per-pair-direction Stage-2 layer-permutation
significance

Per pair-direction \(-\log_{10}(p)\) from the Stage 2 layer-permutation
test (§S2.5). Every observed cluster across the 15 pairwise contrasts
cleared empirical \(p < 0.001\) (i.e., \(-\log_{10}(p) \geq 3\)), the
floor of the empirical null at 1,000 permutations. Cluster mass at the
observed layer ordering is at the upper extreme of the layer-permuted
null in every direction tested. This is reported as a machine-readable
table rather than a figure:
\texttt{code/tfce\_results/llm\_stage2\_perm/stage2\_disc\_p\_values.csv}.

\includegraphics[width=1\textwidth,height=\textheight]{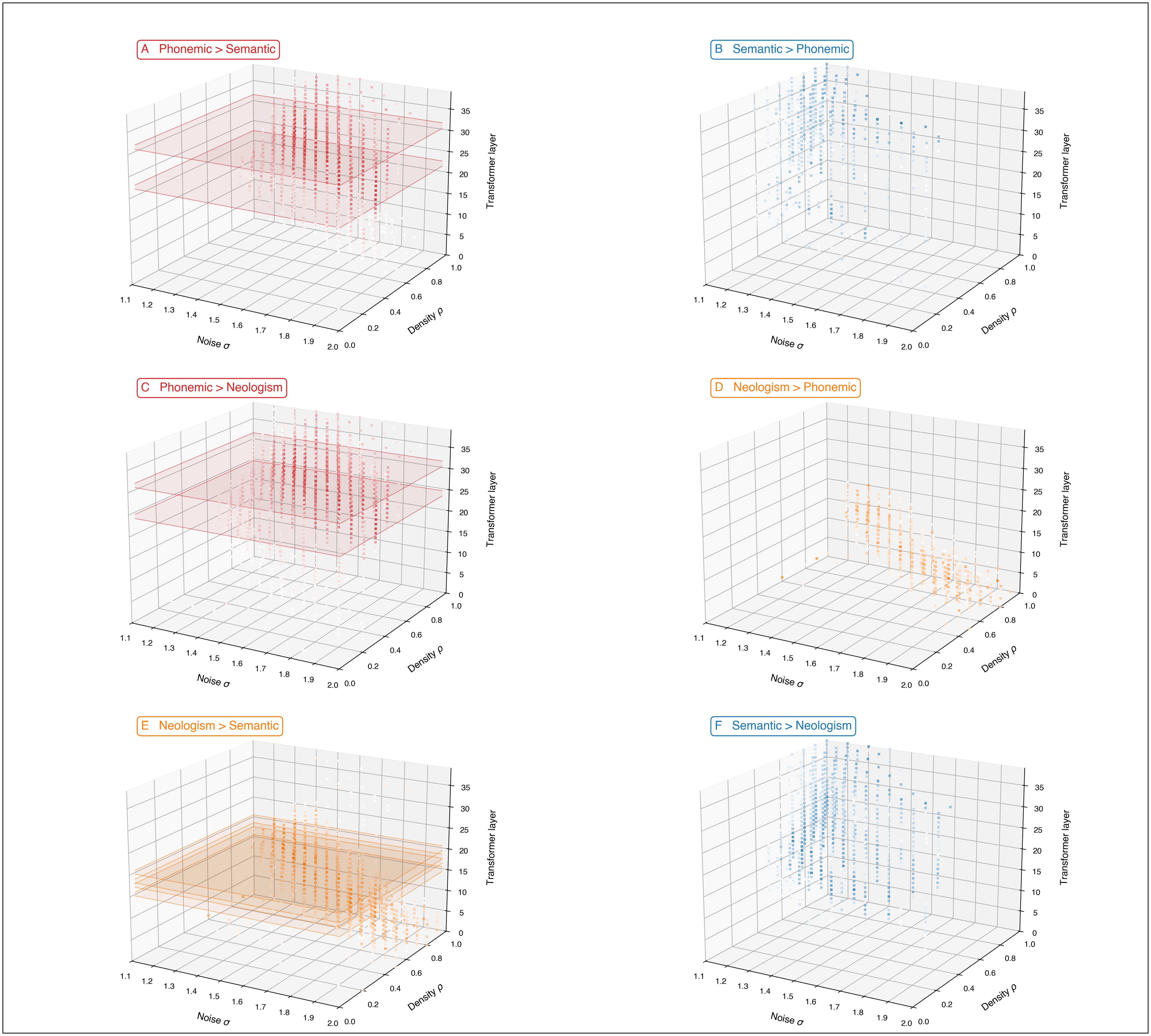}

\textbf{Figure S5.} Descriptive 3D rendering of the (σ × ρ × Layer) data
for the three confirmatory contrasts

Per-cell group-mean \(\bar D(\ell, \sigma, \rho)\) across the 40
discovery seeds, in the panel's directional sign (panel-anchor color
saturates with magnitude). \(\sigma\) (noise standard deviation) and
\(\rho\) (perturbation density) are \textbf{dose parameters and are not
used as spatial axes for cluster correction}, inference is performed
only along the layer axis (Methods §``LLM subtraction analysis''). The
layer-axis TFCE-surviving cluster range from Stage 1 is highlighted as a
translucent slab spanning the full \(\sigma \times \rho\) grid at those
layers, showing that the cluster claim holds across the dose grid rather
than at any specific \((\sigma, \rho)\) cell. This figure is
\textbf{descriptive only} and complements the dose-response analysis
(main-text Figure 5) by showing the full perturbation-grid context for
readers who want it. \textbf{Panel layout}: same A↔F ordering as Figures
2 and 3, \textbf{A} Phonemic \textgreater{} Semantic (primary);
\textbf{B} Semantic \textgreater{} Phonemic (primary); \textbf{C}
Phonemic \textgreater{} Neologism (cross-check); \textbf{D} Neologism
\textgreater{} Phonemic (cross-check); \textbf{E} Neologism
\textgreater{} Semantic (cross-check); \textbf{F} Semantic
\textgreater{} Neologism (cross-check). Panels A, C, and E carry visible
TFCE-surviving slabs; panels B, D, and F do not (consistent with the
asymmetry reported in main text).

\end{document}